# Amplitude-Only FFN Intervention for Tool-Structured LLM Inference: Method, Gated Evaluation Protocol, and Cross-Model Empirical Results

**Sheng Xu (1)**
xusheng.xs@alibaba-inc.com

**Zhen Chen (1)**
moriarty.cz@alibaba-inc.com

**Junhua Wang (1)**
junhua.w@alibaba-inc.com

**Boyuan Huang (1)**
boyuan.hby@alibaba-inc.com

**Ke Jia (1)**
jiake.jk@alibaba-inc.com

**Jiadun Zhu (1)**
jifeng.zjd@taobao.com

**Yiming Xu (2)**
yuzuru.xu@mail.utoronto.ca

*(1) Alibaba Cloud (2) University of Toronto*

## Abstract

We study amplitude-only rescaling of feed-forward network (FFN) activations to repair tool calls and other structured outputs, using multiplicative coordinate masks rather than additive steering vectors. Amplitude Gating (AG) keeps pretrained weights fixed and applies magnitude/entropy-based masks at six SwiGLU sites, with a Newton-Schulz windowed ranking variant. Across three Qwen checkpoints and six text/tool endpoints, effects are task dependent. Under frozen local scorers, Qwen3.5-9B non-NS configurations improve two function-call endpoints by +2.61/+2.67 percentage points (pp) and static ToolBench in/out-domain by +1.64/+2.82 pp; JSON gains are small. These gains survive comparing the weaker of two observed AG runs with the stronger baseline, supporting observed-run robustness rather than population significance. Scores measure lexical or answer/action-name agreement, not executed-tool success. Under the current experimental conditions, the six-task random-forest refit improves native score by +0.46 pp with a positive conditional paired interval; transfer to other settings requires further evaluation. The results support task-specific interventions; causal mechanisms, equal-budget advantages, repeated-run uncertainty, and serving cost require further evaluation. Code and experimental data: https://doi.org/10.5281/zenodo.23654706.

## 1. Introduction

Many activation-steering methods add an intervention vector, $x'=x+\beta v$; Activation Addition (ActAdd) [26] and Contrastive Activation Addition (CAA) [27] derive it from contrasting prompts or examples. We explore multiplicative intervention inside Transformer feed-forward networks (FFNs) [1] to repair tool-structured generation, where incorrect functions or arguments can change the requested action [7].

**Terminology and operator boundary.** Steering is also used broadly for activation interventions, including rescaling [25]. Here we adopt its narrow additive meaning to distinguish direction-adding offsets from magnitude-only coordinate modulation. This convention does not exclude scaling from the broader steering literature. Under SwiGLU's multiplicative interaction [2], the operators have different geometry and potential failure modes: additive offsets can activate a zero coordinate, whereas finite diagonal scaling cannot; nonuniform scaling can nevertheless change the vector's direction. Our focus is adaptive coordinate selection and its repair--regression balance.

Our research logic is **ORP → amplitude multiplication (A) → coordinate-gated AG**. ORP applies an SVD-derived matrix projection that generally mixes coordinates. Its failures motivate restricting the operation to per-coordinate scalar coefficients. G then selects those coefficients using entropy or energy. Scalar coordinate multiplication is itself a diagonal matrix operation, so this distinction concerns operator structure rather than mutually exclusive mathematical classes.

ORP's GSM8K score falls from 87.9454% to 8.4155%, motivating diagonal scaling (Section 3.3). **Coordinate-gated AG is the main empirical direction:** historical P2b/e0.93 produces 34 local action-name repairs and 10 regressions on ToolBench out-domain (Section 3.4). Six SwiGLU sites [2] test shared versus gate-only inputs, gate/value/fused signals, and the output update (Section 3.6).

The site comparison is task dependent: P2b, which derives its mask from the up-projected value signal, gives the largest displayed ToolBench out-domain gain, while shared-input P1 leads on Hermes function call. We hypothesize that P2b offers a useful selection point in SwiGLU: after value-channel expansion, before gate/value fusion and output mixing. Earlier input masking and later fused/output masking may lose useful selectivity. Section 6.1 develops this hypothesis; the observations do not establish a universally best site or a causal explanation.

The corrected September collection covers Qwen3.5-9B, Qwen3-8B [3], and Qwen2.5-7B [4] across six text/tool endpoints. Displayed Qwen3.5 non-NS function-call gains are +2.61/+2.67 percentage points (pp), and ToolBench in/out-domain gains are +1.64/+2.82 pp. These routes retain positive margins across the two observed baseline/AG executions; JSON gains are small. Scorers measure lexical or answer/action-name agreement, not executed-tool success.

Coordinate G is distinct from output selection and acceptance. Protocol-specific offline selector and Per-Sample Fix-Harm Evaluation (PFHE) analyses complement the fixed-route results. Under the current experimental conditions, RF shows a positive conditional effect in the six-task refit; transfer to other settings requires further evaluation. Causal attribution, repeated-run uncertainty, output acceptance, and serving cost remain open. Code and experimental data are archived on Zenodo [28]: https://doi.org/10.5281/zenodo.23654706.

## 2. Related Work

AG sits between several active lines of work, but its goal and operating level are different from each of them.

**FFN sparsification and acceleration.** Fast feed-forward and conditional FFN methods reduce inference cost by executing fewer neurons or replacing dense FFN computation with cheaper conditional computation [12]. AG is not primarily a speed method. It deliberately keeps the pretrained FFN weights and uses amplitude masking as a candidate-generating intervention whose value is judged by native task effects, PFHE, and selector validity. This makes AG complementary to acceleration work: a faster FFN does not by itself decide when a modified activation is safer than the baseline.

**Activation intervention, additive steering, and scaling.** Representation Engineering studies monitoring and controlling model representations [13]. ActAdd derives additive offsets from contrasting prompts [26]; CAA averages contrastive residual-stream differences to obtain behavior-specific offsets [27]. Stoehr et al. learn a scalar for each selected layer/token/site activation vector, with a dynamic variant conditioned on activations [25]. Their scaling is vector-level and can reverse sign; AG instead uses nonnegative coordinate masks derived from current magnitude/entropy or NS history. AG's study concerns this rule, six SwiGLU sites, and paired task effects. Scaling, activation dependence, and fixed weights are not unique to AG, and these methods have not been compared here at matched compute.

**Structured output decoding.** Constrained decoding and JSON-schema-guided generation enforce output syntax at the decoder level [14]. These methods are important for production systems, but they operate after the model has already computed the next-token distribution. AG instead intervenes inside FFN computation before decoding. It therefore targets a different failure mode: cases where internal activation changes can alter which tool, function, or argument structure the model prefers, not only whether the final string obeys a grammar.

**Tool-use and tool-learning systems.** ToolBench and ToolLLM frame tool use as data construction, training, retrieval, and evaluation over API calls [6]. AG does not replace tool training, retrieval, or tool-specific prompting. It is an inference-time candidate mechanism for models already asked to produce tool-structured outputs. Its current evidence is strongest exactly in this setting, but the deployment claim remains conditional on model- and category-specific gates rather than a universal tool-use improvement.

**Candidate selection and abstention.** Random forests combine randomized decision trees to model nonlinear feature interactions [15]. We use this capacity only at the AG control layer: the RF ranks intervention candidates and does not alter the FFN operation itself. A complete controller must also be able to reject every AG candidate and retain the baseline. This requirement is closely related to selective classification, where prediction quality is considered jointly with the coverage induced by abstention [16,17]. Our current RF experiment is intentionally separated from this stronger deployment goal because it evaluates forced candidate selection without a baseline action.

**Matched-pair evaluation.** Aggregate accuracy can hide correlated changes on the same evaluation units. Recent SAT reasoning work introduced a matched-formula protocol and Accurate Differentiation Rate (ADR), which requires jointly correct decisions on minimally different satisfiable and unsatisfiable instances [18]. PFHE shares the principle that paired outcomes reveal behavior that marginal averages can hide, but its pairing axis and estimand are different. ADR pairs two related inputs for one model; PFHE pairs baseline and intervention outputs for the same input and reports the resulting fix, harm, and preservation transitions. The underlying paired 2-by-2 contingency decomposition is standard; our contribution is the intervention-specific evidence contract that couples it to native partial-credit magnitudes, record status, uncertainty, and oracle/selector strata. ADR is therefore methodological context, not a PFHE special case or empirical validation of AG.

## 3. Method: From FFN Intervention to Amplitude Gating

### 3.1 From additive steering to multiplicative intervention

Following the narrow additive convention in Section 1, steering forms $x'=x+\beta v$ [26,27], whereas our FFN interventions multiply the activation: ORP uses a generally non-diagonal projection, and A uses per-coordinate scalar coefficients. A common whole-vector scalar $x'=\gamma x$ is a special case studied in prior activation scaling [25]; coordinatewise coefficients form a diagonal matrix. Magnitude-only describes this coordinate operation, not preservation of the whole vector's direction.

ORP's failure motivates the amplitude restriction; AG then supplies an entropy/energy-based coordinate rule. G selects activation coordinates, separately from sample-level acceptance or baseline fallback. Six SwiGLU sites expose alternative physical signals [2]. Paired outcomes test effectiveness without localizing errors to the FFN.

### 3.2 ORP: matrix projection and its repair hypothesis

Our internal Orthogonal Residual Projection (ORP) applies an SVD-derived projection, $x'=Px$, which generally mixes coordinates. Its hypothesis was that projecting onto an estimated subspace could repair an undesirable output preference. Explored locations included FFN input, internal channels, output, attention, and all-linear projections. The archive calls it Dynamic per-token Orthogonal Residual Projection. Physical-site identity remains essential: historical up-only input intervention is canonical P1v. Section 3.6 gives the canonical mapping; the separate historical overlap below retains its archived checkpoint label.

### 3.3 ORP failures and the need to control fix--harm balance

A historical GSM8K SVD-ORP checkpoint, archived under the label P1/e0.95/w3, is evaluated on a 1,032-question paired overlap: 34 fixes and 56 harms, or 880/1,032 correct (85.27%) versus baseline 902/1,032 (87.40%). Token sequences diverged on 1,012 inputs. This archived-label overlap is reported separately from the corrected full-domain experiment.

The completed September hand-eight-shot GSM8K [8] diagnostic covers 1,319 examples: baseline 87.9454% versus corrected shared-P1 SVD-ORP/e0.95/w3 at 8.4155% (-79.5299 pp). Figure 1 shows this full-domain comparison under the same prompt and scorer.

The binary full-domain means uniquely imply 1,160 baseline-correct and 111 ORP-correct outputs. Consequently, any valid pairing has at least 1,049 harms and at most 111 fixes. These logical bounds establish harm dominance without reconstructing exact paired transition counts.

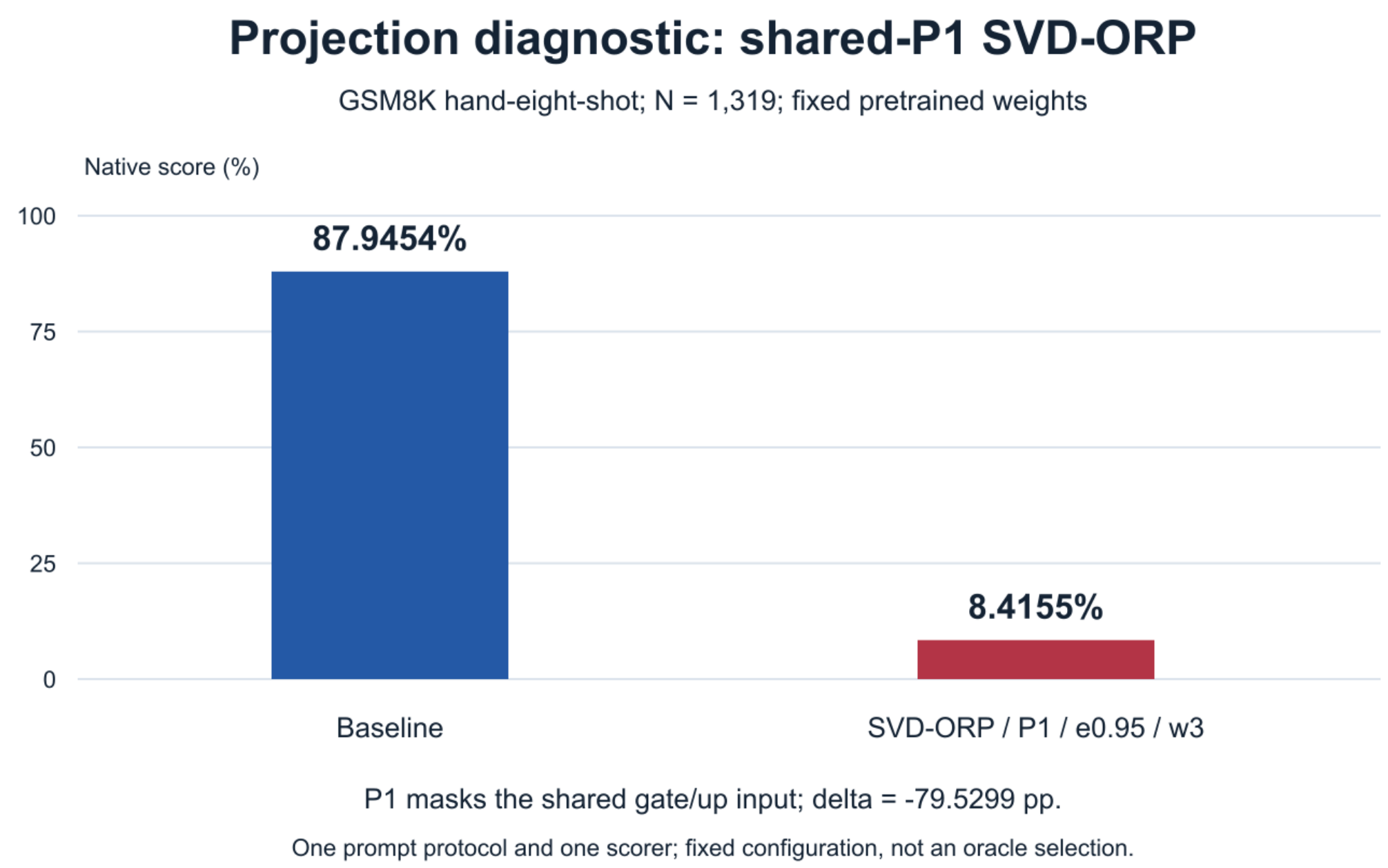

*Figure 1. Full-domain projection diagnostic on hand-eight-shot GSM8K (N=1,319): baseline versus corrected shared-P1 SVD-ORP/e0.95/w3 under the same protocol. Native scores only.*

These failures motivate per-coordinate scalar multiplication, without condemning all projections or establishing A's usefulness. Early entropy-AG at canonical P1v also shows threshold sensitivity: on 1,319 inputs, e=0.93 scores 92.87% with 21 fixes/12 harms; e=0.97 has 11/16, and e=0.80 has 29/31. The archive records 1,137 stable-correct, 54 stable-wrong, and 128 flipping inputs (49 benefit-only, 79 harm-only). These entropy-AG observations illustrate the sensitivity of fix--harm balance to the coordinate rule and retained-mass threshold.

### 3.4 From amplitude multiplication (A) to coordinate-gated AG

**A: the amplitude operation.** A multiplies each activation coordinate by a scalar coefficient; this operator label is separate from the PFHE fix-count symbol. Its interface is

$$x'_{t,i} = \alpha_i(x_t, \text{history}) x_{t,i} \tag{1}$$

**G: selecting activation coordinates.** In the primary experiments, G chooses K and sets `alpha_i=M_K(i)`: retained coordinates keep their values and the rest become zero. Non-NS AG uses magnitude ranking and retained entropy mass; NS changes the ranking using recent activation history (Section 3.7). Energy masks and continuous shrinkage define other archived variants. The main question is whether coordinate selection yields more repairs than regressions.

**Task-specific paired evidence.** Under the historical ToolBench out-domain protocol, fixed non-NS P2b/e0.93 yields 34 normalized action-name-containment fixes and 10 harms over 781 inputs, with a 43.15% baseline. This retrospectively highlighted fixed route shows a favorable task-specific balance. Its paired counts belong to that protocol; Tables 11--11a report the separate September evaluation, whose baseline is 43.4059%.

A binary diagonal mask is also a coordinate projection. Amplitude-only here means nonnegative scaling or masking in the existing basis: uniform positive scaling preserves vector direction, unequal scaling need not. Evaluated coefficients lie between zero and one; amplification above one is a possible extension. Coordinate G is distinct from the prospective output acceptance/fallback controller in Sections 3.8--3.9.

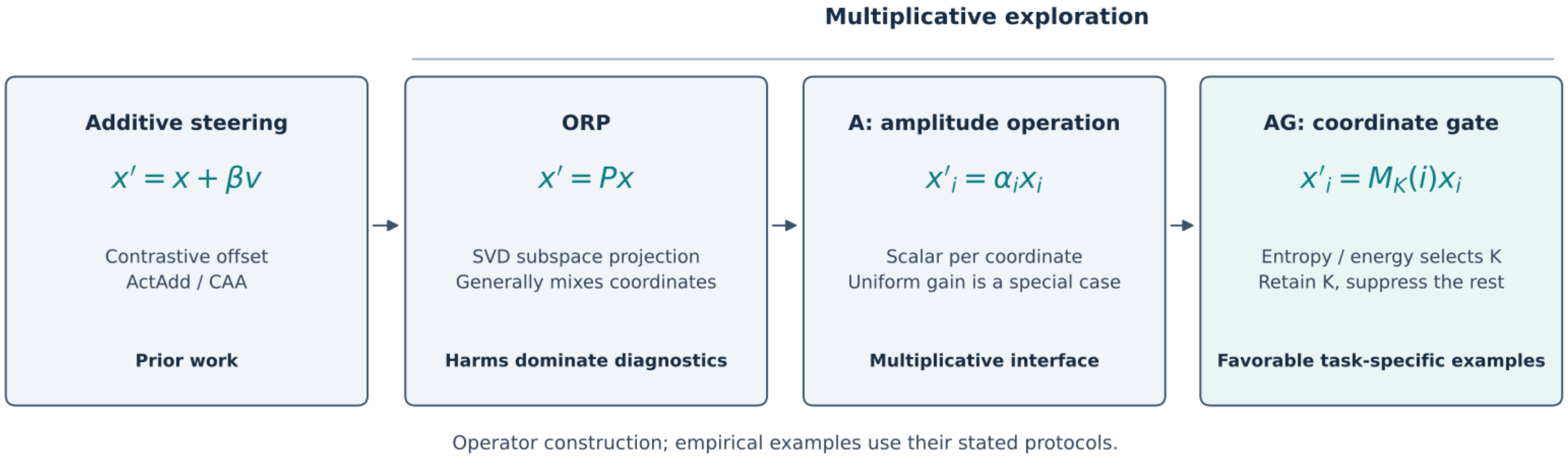

*Figure 2. Research logic from additive steering to multiplicative FFN intervention. ORP applies a generally non-diagonal projection; A denotes per-coordinate scalar multiplication; AG supplies an entropy/energy-based coordinate-selection rule. Arrows describe operator construction; empirical comparisons evaluate complete ORP or AG configurations under their separate protocols.*

### 3.5 FFN computation and the SwiGLU instantiation

For the Qwen checkpoints studied here, the FFN uses separate gate and value branches in a SwiGLU block [2]. With normalized input x, gate activation g, value activation u, fused channels f, and FFN update y, the computation is

$$\begin{gathered} x = \mathrm{Norm}(h), \\ g = \mathrm{SiLU}(W_g x), \qquad u = W_u x, \\ f = g \odot u, \qquad y = W_d f, \qquad h' = h + y \end{gathered} \tag{2}$$

Equivalently, the architectural stages are:

```
h -> Norm(h) -> {gate_proj -> SiLU, up_proj}
        -> elementwise gate/value fusion -> down_proj -> residual addition
```

The normalized input and FFN output have hidden dimensionality; the gate, value, and fused activations have intermediate dimensionality. These spaces expose different coordinates and adaptive selection signals. The six-site design below is the SwiGLU instantiation of AG, rather than a claim of exhaustive coverage of all possible FFN interventions.

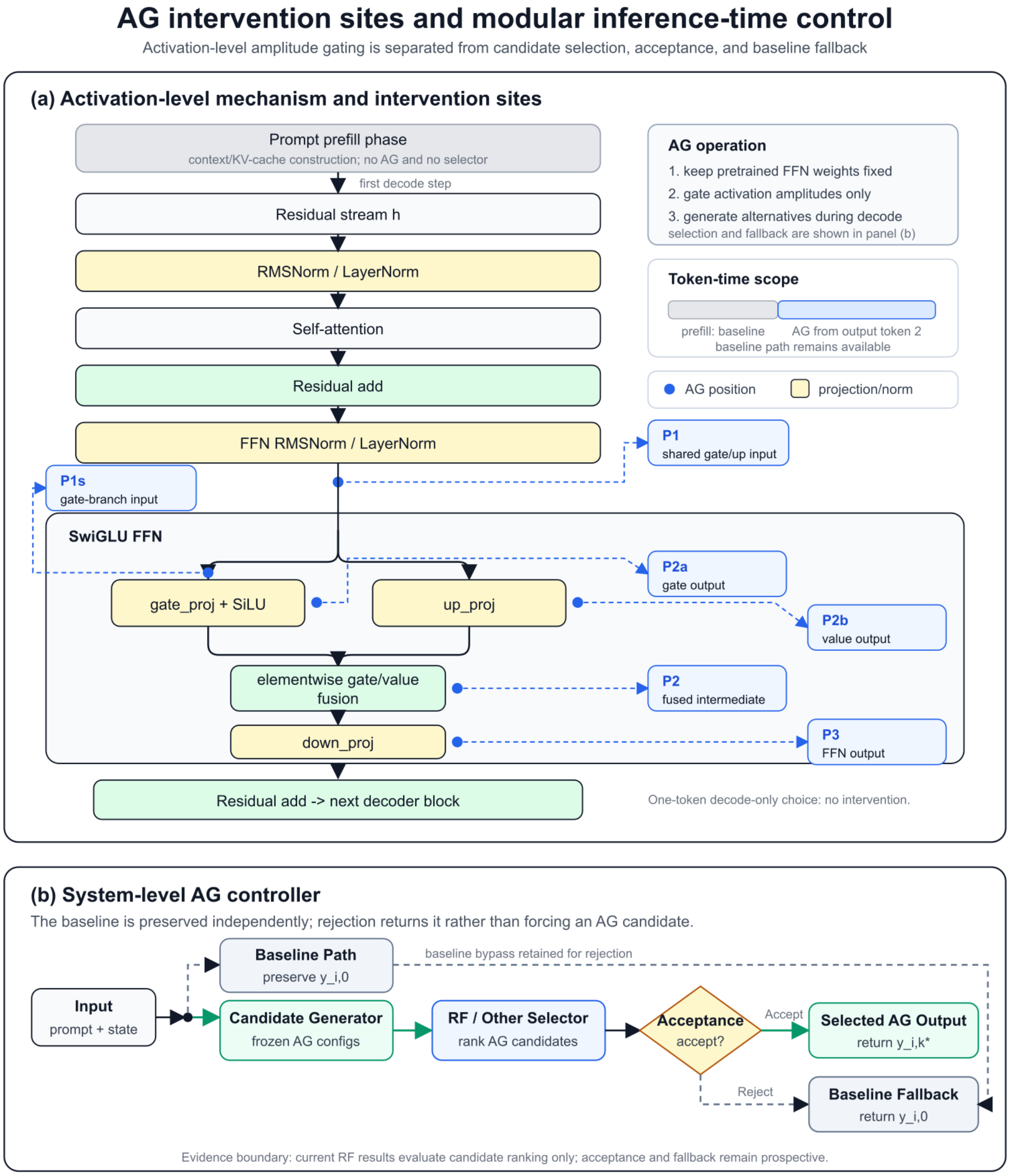

*Figure 3. Two-level AG architecture. (a) A frozen configuration uses one selected site in every FFN layer during decoding. Prompt prefill is unchanged in the text/tool protocol, so the first affected output is generated token 2. (b) Multiple configurations may generate candidates for a selector, followed by a separate acceptance decision and baseline fallback. The current RF experiment evaluates ranking only; acceptance and fallback remain prospective.*

## 3.6 Intervention points

The six sites form three classes (Table 1). They are chosen to compare input coverage, gate/value selection signals, and the FFN update delivered to the residual stream. A configuration uses the same selected site in every FFN layer; it does not combine all six sites in one forward path.

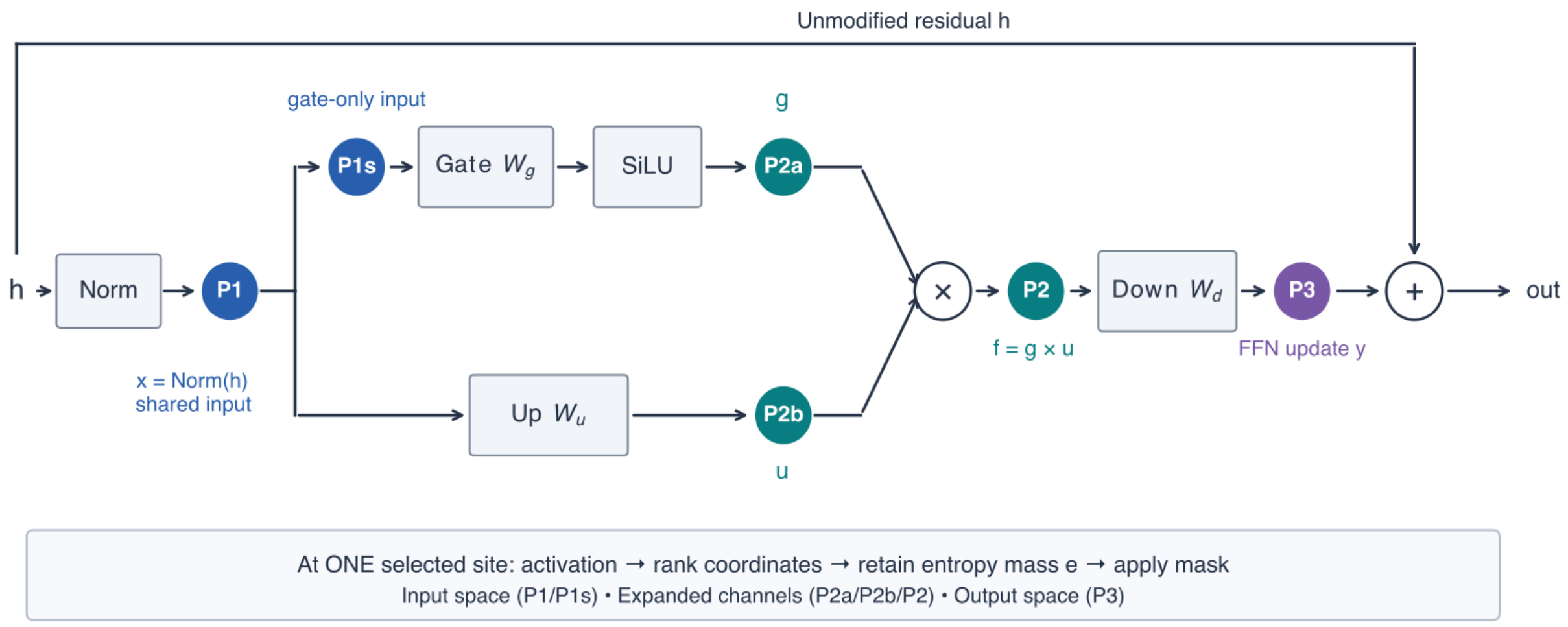

*Figure 4. Six alternative AG sites within a SwiGLU FFN. P1 masks the shared input, whereas P1s masks only the gate-branch input. P2a, P2b, and P2 select coordinates using gate, value, and fused signals; P3 masks the FFN update before residual addition. A configuration uses one selected site in every FFN layer. Weight matrices remain fixed.*

**Table 1. SwiGLU AG intervention sites, physical signals, and comparison roles.**

| Point | Class and location | Dimensionality | Comparison role |
|---|---|---|---|
| P1 | Input: normalized x supplied to both gate and up projections | hidden | Shared-input reference; historical raw p1s |
| P1s | Input: normalized x supplied only to the gate projection | hidden | Gate-only versus shared coverage; newly measured at the true site |
| P2a | Expanded channel: g after gate projection and SiLU | intermediate | Select coordinates from the post-nonlinearity gate signal |
| P2b | Expanded channel: u after up projection | intermediate | Select coordinates from the value signal |
| P2 | Expanded channel: f after gate/value fusion, before down projection | intermediate | Select coordinates from the fused signal |
| P3 | Output: y after down projection, before residual addition | hidden | Modify the FFN update in output coordinates |

**Input-space comparison.** P1 changes the input seen by both branches; P1s changes only the gate-branch input and leaves the value-branch input unchanged. Input projection matrices mix coordinates, so removing one input coordinate can affect many expanded channels. The up-only variant, P1v, is measured separately as a diagnostic control outside the corrected six-site, 24-candidate main pool.

**Expanded-channel comparison.** P2a, P2b, and P2 observe g, u, and f, respectively. Their adaptive rules can select different masks because they use different activations and histories. For the same layer, token, and base forward pass, however, the same fixed coordinatewise multiplier applied once satisfies `(a_j g_j) u_j = g_j (a_j u_j) = a_j f_j`. Thus these labels compare mask-selection signals; they do not by themselves establish three independent suppression mechanisms. The identity holds for binary and continuous multipliers, is algebraic rather than a bitwise floating-point guarantee, and does not extend to non-diagonal SVD projection.

**Output-space comparison.** P3 masks the FFN update after the down projection and leaves the residual bypass h unchanged. Because the down projection mixes expanded channels, masking its output is generally different from masking channels before that projection.

The physical mapping is essential: historical raw `p1s` implemented shared P1, and raw `p1` implemented up-only P1v. The September matrix evaluates genuine gate-only P1s alongside shared P1. Original run keys retain their archived bytes; interpretations use canonical physical names. Dimensionality and location alone do not establish an ordering of quality, safety, or serving cost. We therefore treat all six sites as experimental comparisons rather than prescribe a deployment ranking.

### 3.7 AG taxonomy, formulas, and comparison

The primary text/tool comparison holds the entropy-mass prefix rule fixed and changes its coordinate ordering. Non-NS AG ranks by current absolute activation; NS AG ranks by a short-window Newton-Schulz score. Both select a prefix whose accumulated entropy contribution reaches e of the current activation's total.

Define the current entropy masses:

$$a_i = |x_{t,i}| \qquad p_i = \frac{a_i}{\sum_j a_j + \varepsilon} \qquad m_i = -p_i \log(p_i + \varepsilon) \qquad H = \sum_i m_i \tag{3}$$

Here e is a retained-entropy-mass target, not a retained-coordinate fraction. Given any descending coordinate ordering pi, retain the shortest prefix satisfying the mass target and apply its binary mask. In particular, non-NS entropy AG uses

$$\pi = \mathrm{argsort}(|x_t|, \mathrm{descending}), \quad k_{\mathrm{entropy}} = \min\{k : \sum_{r=1}^{k} m_{\pi_r} \geq eH\}, \quad K_{\mathrm{entropy}} = \{\pi_1, \ldots, \pi_{k_{\mathrm{entropy}}}\}, \quad x'_{t,i} = x_{t,i} M_{K_{\mathrm{entropy}}}(i) \tag{4}$$

The ordering is by absolute activation, not by m_i. Consequently e controls mask size indirectly and does not guarantee preservation of task information.

**NS AG.** NS retains the same current entropy masses but changes the ordering:

$$
\begin{gathered}
W_t = [x_{t-w+1}; \ldots; x_t] \in \mathbb{R}^{w \times d} \\
Y_0 = \frac{W_t}{\|W_t\|_F + \varepsilon} \\
Y_{r+1} = \frac{1}{2}(3Y_r - Y_r Y_r^\top Y_r), \quad r = 0, \ldots, R-1 \\
s_t^{\mathrm{NS}} = \mathrm{abs}(\mathrm{lastrow}(Y_R)) \\
\pi = \mathrm{argsort}(s_t^{\mathrm{NS}}, \mathrm{descending}) \\
k_{\mathrm{NS}} = \\
\min\{k : \sum_{r=1}^{k} m_{\pi_r} \geq eH\} \\
K_{\mathrm{NS}} = \{\pi_1, \ldots, \pi_{k_{\mathrm{NS}}}\} \\
x'_{t,i} = x_{t,i} M_{K_{\mathrm{NS}}}(i)
\end{gathered}
\tag{5}
$$

The displayed W_t is the full-window case. In the evaluated text/tool routes, R=2 and w=4; each sample and layer has its own history, reset at prompt prefill. Early decode steps use the available history without padding. With one activation the implemented score reduces to magnitude ranking, whereas multiple rows introduce local trajectory information. The absolute value is elementwise, so the score has one entry per coordinate. Newton-Schulz ranks activations here: it does not compute entropy, construct an SVD projection, orthogonalize model weights, or update them. This differs from training-time uses such as Muon [5].

The broader AG family includes the alternative operations below. Their formulas are retained to identify the archived variants; they are not interchangeable with the primary entropy masks. In these variants, e denotes retained L2 energy for hard/soft/NMF-style gates, a standard-deviation multiplier for local-norm AG, and retained entropy mass for entropy AG. Top-k uses an explicit coordinate budget k. We retain these archival parameter names in run keys. For any selected set K, M_K(i)=1 if i is in K and zero otherwise.

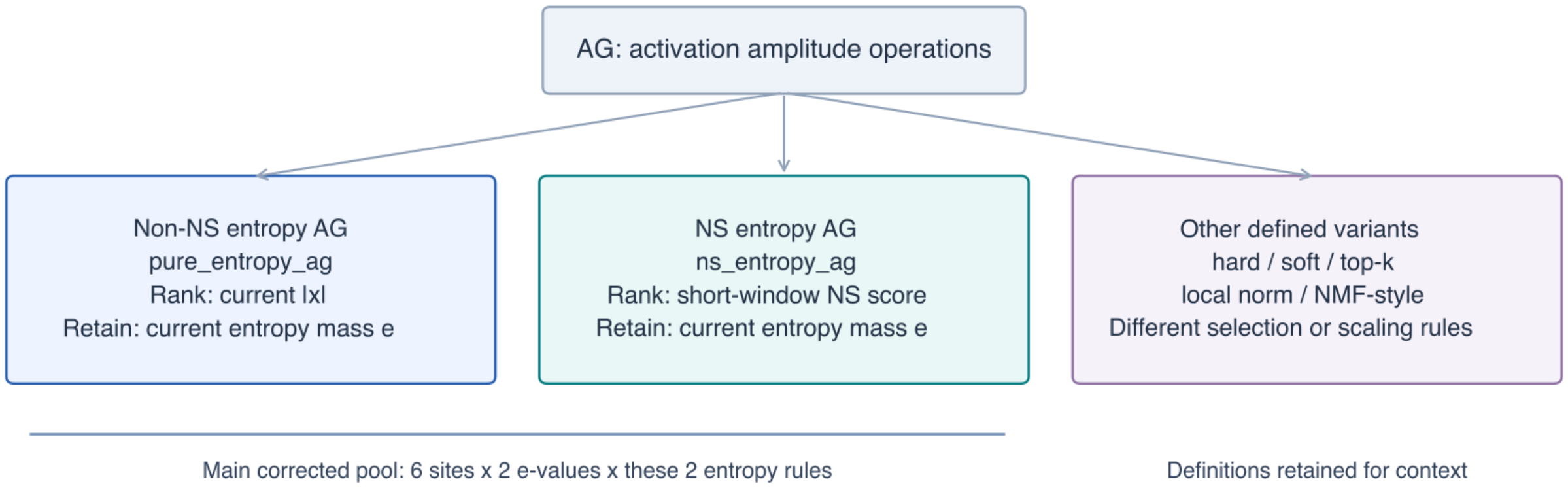

*Figure 5. AG taxonomy: coordinatewise operations include energy masks, shrinkage, fixed top-k, local-norm, and signed energy masks. The primary non-NS/NS comparison uses one entropy-mass mask interface with different coordinate rankings.*

**Hard AG.** Keep the shortest descending-magnitude prefix reaching the L2-energy target:

$$
\begin{gathered}
E = \sum_i x_{t,i}^2 \\
\pi = \mathrm{argsort}(x_{t,i}^2, \text{descending}) \\
k_{\mathrm{hard}} = \\
\min\{k : \sum_{r=1}^{k} x_{t,\pi_r}^2 \geq eE\} \\
K_{\mathrm{hard}} = \{\pi_1, \ldots, \pi_{k_{\mathrm{hard}}}\} \\
x'_{t,i} = x_{t,i} M_{K_{\mathrm{hard}}}(i)
\end{gathered}
\tag{6}
$$

**Soft AG.** Derive a threshold from the same energy prefix, then shrink amplitudes:

$$
\begin{gathered}
\tau = |x_{t,\pi_{k_{\mathrm{hard}}}}| \\
x'_{t,i} = \mathrm{sign}(x_{t,i}) \max(|x_{t,i}| - \tau, 0)
\end{gathered}
\tag{7}
$$

**Top-k AG.** Retain a fixed number of coordinates:

$$
\begin{gathered}
K_{\mathrm{topk}} = \mathrm{TopK}_i(|x_{t,i}|, k) \\
x'_{t,i} = x_{t,i} M_{K_{\mathrm{topk}}}(i)
\end{gathered}
\tag{8}
$$

**Local-norm AG.** Threshold absolute activations by their standard deviation, without sorting:

$$\sigma_t = \mathrm{std}(x_t)$$

$$K_{\mathrm{local}} = \{i : |x_{t,i}| > e\sigma_t\} \tag{9}$$

$$x'_{t,i} = x_{t,i} M_{K_{\mathrm{local}}}(i)$$

**NMF-style AG.** Apply separate L2-energy prefixes to positive and negative components, then recombine:

$$\begin{gathered}
x_t^+ = \max(x_t, 0), \qquad x_t^- = \max(-x_t, 0) \\
E_v = \sum_i v_i^2, \\
\pi^v = \mathrm{argsort}(v_i^2, \mathrm{descending}) \\
k_v = \min\{k : \sum_{r=1}^{k} v_{\pi_r^v}^2 \geq eE_v\} \\
K_v = \{\pi_1^v, \ldots, \pi_{k_v}^v\} \\
[g(v, e)]_i = v_i M_{K_v}(i) \\
x'_t = g(x_t^+, e) - g(x_t^-, e)
\end{gathered} \tag{10}$$

**Table 2. AG variants and their selection signals.**

| Variant | Selection signal | Operation | State/window | Role |
|---|---|---|---|---|
| **pure_entropy_ag** | Current magnitude ranking; entropy-mass prefix | Binary mask | Stateless | Main non-NS text/tool candidate |
| **ns_entropy_ag** | NS-window ranking; current entropy-mass prefix | Binary mask | Up to four decode activations | Main NS text/tool candidate |
| **hard** | L2-energy prefix | Binary mask | Stateless | Alternative energy gate |
| **soft** | L2-derived tau | Amplitude shrinkage | Stateless | Alternative continuous operation |
| **topk** | Fixed top-k magnitude | Binary mask | Stateless | Explicit coordinate budget |
| **local_norm** | Local standard-deviation threshold | Binary mask | Stateless | Scale-adaptive threshold |
| **nmf** | Positive/negative L2 prefixes | Signed recombination | Stateless | Separate signed energy selection |

### 3.8 Modular AG control pipeline

Applying AG and deciding to use its output are distinct operations. A fixed-route experiment uses one frozen configuration for every input in its declared scope and generates one AG output per input. It does not rank several completed outputs for each sample. A multi-candidate experiment instead produces a pool that a selector may rank.

For evaluation unit i, let $y_{i,0}$ denote the baseline output and $C_{AG}=\{1,...,K\}$ the frozen AG configuration set. Candidate k produces $y_{i,k}$ and feature vector $z_{i,k}$. A complete controller has four stages:

1. **Generate:** obtain outputs from the frozen AG configurations. 2. **Rank:** assign candidate utilities using features available at the declared decision time. 3. **Accept or abstain:** compare the selected candidate with a pre-specified acceptance condition. 4. **Fall back:** return the independently preserved baseline when abstaining.

The baseline is excluded from the archived AG ranking and combination-oracle sets so that their scores measure alternatives supplied by AG. It remains a separate action in the proposed controller. A combination oracle sees evaluation scores and measures headroom; a learned selector estimates ranking quality on held-out units; neither alone validates acceptance or fallback. Features requiring completed trajectories, evaluator outputs, or correctness labels cannot substantiate a decision claimed to occur before they are available. The current selectors force an AG choice, so their results remain exploratory offline evidence.

The text/tool experiments use decode-only AG. Prompt prefill is unchanged; generated token 1 is selected from prefill logits, and the first AG decode step can affect generated token 2. For a multi-token prompt and a one-token output, the verified guard therefore yields baseline-equivalent predictions.

Historical C-Eval/MMLU-Pro and the corresponding HLE choice diagnostics instead enable last-token prefill intervention. They are reported as a separate prefill-on protocol.

### 3.9 Random-forest candidate selection within AG

We instantiate the candidate-ranking stage with a random forest (RF) as an exploratory nonlinear selector. Let `r_phi(z_{i,k})` be the forest score assigned to candidate `k`. Candidate ranking selects

$$\hat{k}_i = \arg\max_{k \in \mathcal{C}_{\mathrm{AG}}} r_\phi(z_{i,k}) \tag{11}$$

A complete controller additionally learns or specifies an acceptance function `a_phi`. Its returned output is

$$\begin{aligned} y_i &= y_{i,\hat{k}_i} \quad \text{if} \quad a_\phi(z_{i,\hat{k}_i}) = 1, \\ y_i &= y_{i,0} \quad \text{otherwise} \end{aligned} \tag{12}$$

The distinction between `r_phi` and `a_phi` is essential. Ranking asks which AG candidate is preferable *among candidates*; acceptance asks whether that candidate is preferable to retaining the baseline. A system that implements only the first operation is forced to intervene even when every AG candidate is harmful.

RF does not modify the AG mask, intervention position, retained-mass parameter, or pretrained model. It is one possible selector over the candidate space produced by those choices. Its appeal is its ability to model nonlinear feature interactions without imposing a linear utility surface. However, every selector feature must be labeled by decision time. Request-level, prefill-level, and legally available early-generation features may support an online policy; features requiring completed candidate trajectories, evaluator outputs, or answer correctness remain diagnostic-only.

The current RF study evaluates only `r_phi` under forced candidate choice. It does not implement the acceptance function above, train on an explicit baseline action, or validate a prospective risk--coverage policy. We therefore describe RF as an exploratory candidate selector within the modular AG framework. The six-task RF refit shows a positive conditional ranking effect; a stable benefit across partitions and a deployment-ready acceptance gate require further evaluation.

# 4. Experiment Design: Coverage and Credibility Protocol

## 4.1 Design goals: coverage without overclaiming

The experiment design has to satisfy two competing requirements. It must be broad enough to avoid optimizing for one accidental model, dataset, position, or e-value, but constrained enough that the resulting evidence can still be interpreted. We therefore separate two questions throughout the protocol:

1. Does the AG candidate space contain useful alternatives? 2. Can a fixed configuration or online-feature gate choose those alternatives without seeing the answer score?

The first question is answered by oracle and candidate-space diagnostics. The second requires train-selected fixed configurations or held-out selectors, and a deployment claim additionally requires legal feature timing, acceptance, and fallback. This distinction is the main protection against overclaiming.

**Table 3. Experimental coverage axes and the claim each axis is designed to test.**

| Axis | Coverage choice | Why it matters |
|---|---|---|
| Model architecture | Qwen3.5-9B hybrid linear/full attention; Qwen3-8B and Qwen2.5-7B dense full attention | Tests whether FFN-level AG remains relevant when attention structure changes |
| Task category | Tool-structured and financial QA, plus separately reported choice diagnostics | Prevents a tool-only result from being mistaken for a universal text result |
| FFN position | P1/P2/P3 inherited from ORP; P1s/P2a/P2b added from SwiGLU structure; here P means position/site, not a statistical p-value | Separates input, gate-branch, value-branch, fused, and output-side effects |
| AG configuration | Non-NS entropy variants, Newton-Schulz entropy variants, e-values such as 0.93/0.95, and windowed settings; for entropy AG, e is the retained entropy-mass target defined in Section 3.7 | Measures sensitivity to amplitude threshold and local entropy ranking definition |
| Selector feature family | Prefix/format, early logprob trajectory, baseline uncertainty, divergence/energy, and diagnostic-only post-fork features | Separates proposed decision-time signals from explanatory post-fork diagnostics |

This coverage design is why the paper reports both positive and negative evidence. Qwen3.5-9B provides the only baseline-aligned old-P/new-P learned- selector result. Qwen3-8B and Qwen2.5-7B expose a more fundamental boundary: their archived sources cannot be pooled into one 24-candidate protocol, so the paper uses only source-separated fixed-route and oracle summaries for those models. This prevents a broad sweep from being mistaken for a universal switch.

## 4.2 Candidate matrix and intervention coverage

For each model and dataset, we evaluate an explicit baseline and one or more AG candidate matrices. The September main collection covers six tasks on each of three models: 18 model/task scopes, 522 configurations, and 773,082 evaluation rows. Each scope contains a fresh baseline, 24 candidates at canonical P1/P1s/P2/P2a/P2b/P3, and four separate P1v diagnostic candidates. Supplementary GSM8K and choice diagnostics use their separately declared protocols.

Results come from two separately defined collections. The September corrected-site collection supplies fixed-route scores and decode-only controls. Offline selector, PFHE-format, prefill-on choice, feature, and post-hoc gate analyses use the archived matrices and selected-output records under the sample universes, baselines, scorers, and splits specified below. These completed measurements retain their own protocols; scores and paired intervals are not pooled across collections. Fixed routes, learned selectors, and label-informed combination oracles are reported separately. RF and ConvFinQA selectors are refitted on their declared six-task and single-task populations, using the retained measurement matrices.

Within the archived old-P/new-P sources, Qwen3.5-9B has a source-aligned 24-candidate union: 12 old-P and 12 new-P candidates for 8,886 units across the six tasks, with consistent paired baseline scores and correctness. The Qwen3-8B and Qwen2.5-7B sources use distinct sample domains or baseline outcomes and are analyzed separately. In canonical physical names, the archived old-P set covers P1v/P2/P3 and the archived new-P set covers P1/P2a/P2b; both include non-NS and Newton-Schulz entropy variants at retained-mass settings 0.93 and 0.95.

The notation is deliberately literal. `P` denotes the FFN intervention position, not statistical significance. In run keys such as `p1s` or `p2b`, it tells us the archived identifier. Its physical meaning must be resolved through the site mapping above rather than inferred from the raw string. For the entropy candidates emphasized in the text/tool results, `e` denotes the retained entropy mass used to choose the mask prefix. Thus e0.93 and e0.95 are two retained-mass settings at the same FFN point. Each aligned old-P and new-P block is therefore a matched candidate design: three positions crossed with two e-values and two AG families (non-NS entropy AG and NS entropy AG, with `w4` denoting the NS window). This matched design is what allows the P-by-e comparison reported in Section 5 and analyzed in Section 6.1.

The e-values in the matched text/tool tables are not learned from the reported test labels. They are fixed before the route-level result comparison as a small retained-mass grid inherited from the ORP/early-AG sweep: earlier diagnostics showed that nearby retained-mass settings could change fix/harm balance, with values around 0.93--0.95 repeatedly appearing in the sensitive region. We therefore keep e0.93 and e0.95 as paired probes rather than selecting a new best e per model or per dataset. When a table says that a cell such as P2b/e0.93 is strongest, that is a result within this pre-specified grid, not a claim that e=0.93 is globally optimal.

We use two discrete e-values rather than a dense sweep for three practical reasons. First, the purpose of the matched matrix is interaction testing, not hyperparameter optimization: the question is whether position changes the effect at nearby retained-mass settings. Second, every additional e-value multiplies the candidate count by model, dataset, position, and AG family, increasing compute cost and the risk of post-hoc best-cell overinterpretation. Third, 0.93 and 0.95 form a compact bracket inside the previously observed sensitive region: close enough to test local stability, but separated enough to expose sign or fix/harm changes. A denser e sweep remains useful future work, especially for deployment calibration, but it is not used as the main evidence level in this paper.

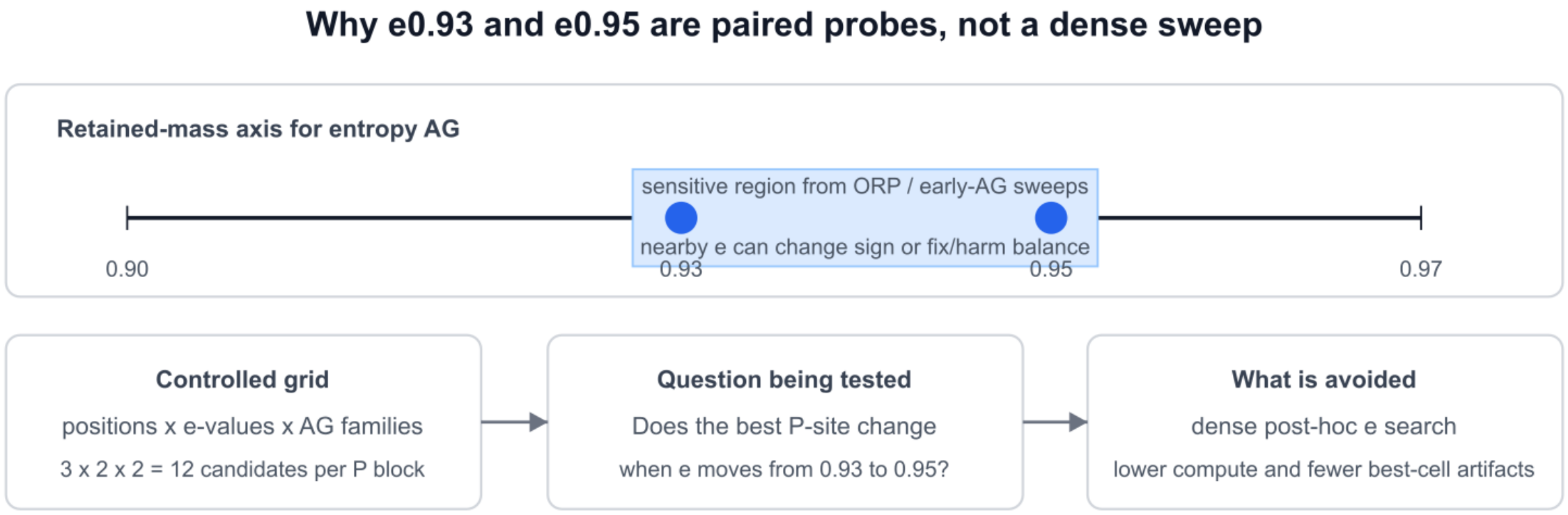

*Figure 6. Matched e-grid design: e0.93 and e0.95 are paired probes inside the sensitive retained-mass region, so the experiment tests P-by-e interaction without turning the candidate matrix into a dense post-hoc e search.*

The candidate matrix is intentionally balanced across AG morphology. For each position/e setting, the non-NS form ranks dimensions by the current activation magnitude, while the NS form ranks dimensions by a Newton-Schulz score. The windowed NS setting, written as `w4` in the run keys, uses the most recent four decode-token activations to estimate a local score for the current token. This design lets the experiment answer two separate questions: whether simple magnitude-ranked entropy AG is already sufficient, and whether local Newton-Schulz scoring adds useful candidate diversity. We do not treat NS as a strictly stronger replacement for non-NS AG; it is a parallel morphology whose value must be established by fixed-configuration, oracle-selection, and learned-selector evidence.

Baseline output is always retained as the evaluation reference. It is not eligible for best fixed single-configuration, combination oracle, or learned-selector candidate choices. This convention makes oracle delta interpretable: oracle results measure whether non-baseline AG candidates contain useful alternatives, not whether an oracle can simply choose the baseline when AG is harmful.

The feature matrices record candidate outputs and proposed gate features. The six-task Qwen3.5-9B union contains 213,264 aligned candidate rows. Qwen3-8B and Qwen2.5-7B measurements retain their source-specific sample domains and baseline protocols. Learned-selector claims are restricted to the aligned Qwen3.5-9B tool category.

## 4.3 Models, datasets, and evaluation routes

The current evidence base covers three concrete checkpoints over six text/tool datasets. We report checkpoint names rather than only family names because the architectural differences matter for interpreting generalization.

**Table 4. Evaluated checkpoints, architecture characteristics, and evidence role.**

| Short label | Checkpoint | Attention structure | FFN shape | Context config | Role in this paper |
|---|---|---|---|---|---|
| Qwen3.5 | Qwen3.5-9B | Hybrid text stack: 32 layers with 24 linear-attention layers and 8 full-attention layers; full attention every fourth layer | SwiGLU-style gated FFN; hidden 4096, intermediate 12288, SiLU | max position 262144 | Strongest function-call/tool-use evidence |
| Qwen3 | Qwen3-8B | Dense full-attention causal LM; 36 layers | SwiGLU-style gated FFN; hidden 4096, intermediate 12288, SiLU | max position 40960 | Historical fixed/oracle boundary; September raw coverage complete |
| Qwen2.5 | Qwen2.5-7B | Dense full-attention causal LM; 28 layers | SwiGLU-style gated FFN; hidden 3584, intermediate 18944, SiLU | max position 131072 | Historical fixed/oracle boundary; September raw coverage complete |

This model set gives a useful but bounded architecture comparison. Qwen3.5-9B tests AG in a hybrid linear/full-attention architecture, while Qwen3-8B and Qwen2.5-7B test AG in dense full-attention architectures with different depth and FFN expansion sizes. All three use SwiGLU-style gated FFNs, which allows the same six-point taxonomy to be evaluated across models. The broader AG principle is not restricted to SwiGLU, however: it requires only identifiable FFN activation sites whose amplitudes can be modulated without changing weight directions. The empirical claim is therefore conservative. AG produces a positive exploratory offline learned-selector result on the source-aligned Qwen3.5-9B union. The other checkpoints show candidate-space headroom only after their old-P and new-P protocols are separated, and their train-selected fixed tool routes are non-positive in native score. This establishes a measured cross-checkpoint boundary while rejecting universal gate transfer.

The six main tasks form two evaluation routes:

- tool_structured_agentic: ToolBench in-domain and out-domain [6], Hermes function call, Hermes function call single-turn, and Hermes JSON mode single-turn [22].
- financial_qa: ConvFinQA dev [21].

The evidence base used in this paper is summarized below.

**Table 5. Evidence inventory and the manuscript claim supported by each source block.**

| Evidence block | Source scope | Used for |
|---|---|---|
| Historical projection and entropy-AG diagnostics | Archived paired diagnostics; full-domain hand-eight-shot GSM8K SVD-ORP comparison | Operation-level motivation and retained-mass sensitivity |
| September corrected-site collection | 18 main model/task scopes; 522 configurations; 773,082 rows | Fixed native scores and gate-only controls; supplements reported separately |
| Qwen3.5-9B aligned mining | Six tasks; old-P + new-P; 213,264 candidate rows | Tool-category exploratory selector evidence |
| Qwen3-8B source-separated mining | Distinct old-P and new-P sample domains and baselines | Fixed-route/oracle boundary; no pooled learned result |
| Qwen2.5-7B source-separated mining | Two 12-candidate sources with distinct baseline outcomes | Fixed-route/oracle boundary; no pooled learned result |
| Qwen3.5 tool feature ablation and bootstrap | Stable hash split over 7,396 tool samples | Feature-family evidence within the aligned tool category |
| Qwen3.5 MC/choice diagnostics | C-Eval and MMLU-Pro rerun records with the stable hash split | Negative evidence against universal choice-task gains |
| Divergence/layer trace diagnostics | Aligned text pools and Qwen3.5 choice trace details | Trajectory diagnostics and limits of layer-trace gating |
| Qwen3 versus Qwen2.5 comparison | Source-specific full-attention protocols | Boundary of cross-checkpoint transfer |
| PFHE-format paired refresh | Frozen selected-output records keyed by evaluation unit; paired bootstrap | Fix/harm, preservation, and native-score magnitude |
| Six-task RF refit | 8,886 units; 12 candidates; five grouped folds | Paired native and strict RF--linear comparison |
| ConvFinQA selector refit | 957 train / 533 test units; 24 candidates | Single-task financial-QA selector results |

The reported analyses specify source hashes, sample universes, split rules, selected run keys, bootstrap procedures, and PFHE-format identity checks. PFHE-format results characterize paired saved records under the stated scoring protocols; prospective operational endpoints are addressed in Section 9.

### 4.4 Gate protocol and leakage controls

Coverage alone is not enough. A broad candidate sweep can easily produce misleading conclusions if the final result is selected after seeing labels. We therefore impose leakage controls before reporting results.

First, every run is compared against an explicit baseline for the same model, dataset, prompt/evaluator, and decoding setup. This prevents an AG result from being compared against a stale or mismatched baseline.

Second, candidate generation and selector evaluation are separated. The candidate matrix may be broad for discovery, but any prospective claim requires held-out selection, decision-time-legal features, an explicit baseline action, and an acceptance rule.

Third, proposed gate features exclude explicit labels and post-generation diagnostic fields. Nevertheless, some retained baseline/candidate trajectory features still require a frozen decision-time and serving-cost contract. The current selectors also force one AG candidate and do not execute baseline fallback. We therefore label their results exploratory offline rather than deployable-feature evidence.

Fourth, the design includes task-category and cross-model stress tests. A result that holds only under a pooled all-dataset view is not considered sufficient. The method must show where it works, where it weakly works, and where protocols or selectors fail.

The current learned-selector result is restricted to the aligned Qwen3.5 protocol and should not be read as a prospective online A/B validation. Repeated grouped splits, time splits, a real fallback action, and new held-out datasets remain necessary before stronger claims can be made.

### 4.5 Accounting levels and gate terminology

We distinguish four evaluation levels.

**Table 6. Evidence strata for AG candidate-space, fixed-route, selector, and diagnostic results.**

| Level | Meaning | Deployment relevance |
|---|---|---|
| Combination oracle | Per sample, choose the best non-baseline AG candidate after observing scores | Upper bound only |
| Train-selected fixed configuration | One AG run key selected on the train split, then applied uniformly to held-out units | Fixed-route evidence; not a per-sample oracle |
| Learned selector | Offline ranker chooses among AG candidates on held-out units | Exploratory while timing and baseline acceptance remain unresolved |
| Diagnostic-only analysis | Uses post-hoc or label-adjacent signals | Explanation only |

We retain the artifact names in tables but interpret them as selectors. A **mixed gate** is a category-level learned ranker that combines multiple online-observable feature families such as prefix/format features, early logprob trajectory, uncertainty, and divergence/energy features. A **ridge gate** is a regularized linear utility ranker over the same proposed feature pool. A **single indicator** gate uses one proposed feature or indicator. These are forced-AG offline candidate rankers; they are neither prospective online A/B results nor complete controllers with fallback.

The standard term in this paper is **combination oracle**. When the candidate set is described as the union of several position subsets, we sometimes refer to the same upper-bound construction as a union over those candidates; it is still a combination oracle and remains non-deployable because labels are observed before choosing the candidate.

We explicitly avoid three forms of overstatement.

Pooled multi-config bias occurs when many AG configurations are mixed and a sample is counted as captured if any configuration fixes it. This is useful for finding headroom but does not correspond to a single online policy.

Best-selected bias occurs when multiple accepted candidates exist and the correct one is chosen after the fact. This is oracle behavior, not deployment.

Row-level bias occurs when sample-by-configuration rows are counted as if they were independent questions. We report sample-level results for deployment claims.

### 4.6 Classical task metrics and statistical protocol

Frozen local evaluation scores are the primary effectiveness endpoints. Hermes function-call and JSON tasks use normalized token-overlap F1. ConvFinQA uses normalized answer-token containment. ToolBench is a static first-Action-name task scored by normalized tool-name containment. Normalization lowercases text, removes English articles and punctuation, and collapses whitespace; multiple references use the maximum score. These endpoints measure lexical agreement or containment, not schema validity, function execution, end-to-end agent success, or human-judged factuality. Archived tables call these local protocol scores native scores. Choice diagnostics use their separately declared accuracy scorers. Let `s_i^B` and `s_i^I` denote scores in [0,1] on the same paired unit. Means are displayed as percentages and differences as percentage points (100 times the paired effect below):

$$\Delta_{\text{native}} = \frac{1}{N}\sum_{i=1}^{N}(s_i^I - s_i^B) \tag{13}$$

For partial-credit tasks, we additionally report the median paired difference and the numbers of positive, negative, and tied score differences. These magnitude summaries remain distinct from strict correctness transitions. A native-score increase is not automatically a strict fix, and a decrease is not automatically a strict harm.

The offline paired analyses use nonparametric paired bootstrap resampling of complete evaluation units with 10,000 replicates and a deterministic SHA-256-derived seed for each result-group key. Category aggregates span heterogeneous scorers and are therefore treated as descriptive route summaries rather than a substitute for dataset-stratified inference. Every analysis records the paired domain; operational evaluations should additionally report requested, successful, paired, timeout, parse-error, and scorer-error counts. Independent resampling of baseline and intervention outputs is not permitted.

We report fixed configurations, held-out learned selectors, and combination oracles separately. A fixed configuration asks whether one frozen AG route improves the task uniformly. A learned selector estimates offline selection quality under its declared feature and split contract. A combination oracle observes scores before choosing a candidate and is only a candidate-space upper bound. These quantities are not entries in a common deployment leaderboard. Scores from heterogeneous binary and partial-credit scorers are not collapsed into an undefined universal endpoint; the mixed strict aggregate at the start of Section 5 remains a candidate-space sanity check only.

**Baseline repeatability and observed-run sensitivity.** A baseline is an observed experimental outcome, not an exact constant. We retained historical baselines and collected fresh baselines with the September candidate runs. We also performed a controlled baseline repeatability diagnostic on a fixed 128-example Qwen3.5 subset, preserving the resident model, raw inputs, batch size 16, and greedy generation settings. Two natural-SDPA executions obtained native scores of 25.1983% and 25.3983%, a 0.2000 pp difference: 122/128 predictions and 124/128 per-example scores agreed. Two executions with cuDNN SDPA disabled, and two under the global strict numerical condition, each agreed on all 128 predictions and scores within their respective condition. These six runs are numerical diagnostics, not six full benchmark replications; cross-condition differences are not counted as within-condition fluctuation, and the probe does not establish the cause of every historical discrepancy.

Historical and September baseline means are compared within matching model/task protocols. These comparisons describe observed execution variation; they are not exchangeable repetitions of one experiment. The six Qwen3.5 main-task baseline comparisons below have a maximum absolute difference of 0.3778 pp. Supplementary GSM8K 89.39% and hand-eight-shot 87.9454% belong to distinct prompt protocols and are not treated as matched repetitions.

For each previously reported, physically matched fixed route, let `B_h,B_s` and `I_h,I_s` denote historical and September baseline and AG native means. We report both within-run effects and the conservative observed-run margin `L_obs = min(I_h,I_s) - max(B_h,B_s)`, in percentage points. A positive margin means that even the lower observed AG mean exceeds the higher observed baseline mean, accounting for the observed variation of both paths. We keep the originally reported configurations and canonical physical-site mapping; we do not select new winners. This margin is a descriptive sensitivity check, not a confidence interval, p-value, or bound on future runs. Formal significance of corrected fixed-route effects requires paired records and run-level uncertainty; the historical selector intervals do not supply that test and are left unchanged.

### 4.7 PFHE: Per-Sample Fix-Harm Evaluation

Aggregate task metrics answer whether an intervention changes average performance, but they do not identify which baseline successes are preserved or destroyed. We therefore introduce **Per-Sample Fix-Harm Evaluation (PFHE)** as a companion paired reporting protocol. PFHE is not a new scalar objective, an oracle, or a routing algorithm. It is a contract for retaining and reporting the joint baseline--intervention outcome on every evaluation unit.

**Paired data contract.** Each pair is keyed by `dataset::sample_id`, not by a bare identifier that may collide across datasets. Baseline and intervention records must share the same prompt, scorer and scorer version, requested sample universe, and score direction. For multiple AG candidates, one row is retained per `(dataset::sample_id, run_key)`, and the result is labeled as fixed, learned-selector, oracle, or diagnostic evidence before aggregation. Pairing is performed before any mean, count, or rate is computed.

For an admissible binary-correctness pair, let `y_i^B,y_i^I in {0,1}` denote baseline and intervention correctness. PFHE partitions the paired sample universe into four mutually exclusive states.

**Table 7. PFHE four-state partition for paired baseline and intervention correctness.**

| PFHE state | Baseline | Intervention | Count | Interpretation |
|---|---|---|---|---|
| Fix | 0 | 1 | `A` | AG corrects a baseline error |
| Harm | 1 | 0 | `H` | AG breaks a baseline success |
| Preserved correct | 1 | 1 | `C` | Both paths are correct |
| Preserved wrong | 0 | 0 | `D` | Neither path is correct |

The decomposition satisfies

$$N = A + H + C + D,$$

$$\mathrm{Acc}_B = \frac{H + C}{N}, \tag{14}$$

$$\mathrm{Acc}_I = \frac{A + C}{N},$$

and hence

$$\Delta\mathrm{Acc} = \mathrm{Acc}_I - \mathrm{Acc}_B$$

$$= \frac{A - H}{N} \tag{15}$$

Each PFHE-compliant result retains all four counts and their common denominator `N`; rates are count divided by `N`. We also report correctness-transition coverage `(A+H)/N`, baseline-error recovery `A/(A+D)`, regression risk among baseline successes `H/(H+C)`, and the changed-case fix share `A/(A+H)` as a descriptive quantity. Undefined denominators remain undefined. The discordant counts support exact paired tests such as McNemar's test [19], while interval estimates use paired resampling [20].

**Partial-credit tasks.** Partial credit is a magnitude attribute, not a fifth PFHE state. We report paired mean and median native-score differences, positive/negative/tied counts, magnitude distributions, uncertainty, and record status. Strict `A/H/C/D` is added only when the task has a defensible full-credit rule. A raw score change of 0.1 and a crossing between score deciles are different events and must not share a label. Historical `partial10` results are interpreted under their respective raw-change or decile-crossing definitions and kept separate from the confirmatory PFHE analysis.

PFHE may accompany a classical result only when both views use the same paired domain, baseline, selected output, scorer, candidate pool, and split. We do not combine a classical score from one protocol with transition counts from another, and we do not reconstruct the joint table from marginal accuracies. This same-output requirement distinguishes PFHE from Accurate Differentiation Rate (ADR) [18]: ADR evaluates one model across two minimally different SAT formulas, whereas PFHE evaluates baseline and intervention on the same input. The shared principle is paired reporting; neither metric is a special case or empirical validation of the other.

### 4.8 Feature policy and evidence boundary

The mining reports separate gate-policy features from diagnostic-only features.

**Table 8. Selector feature timing and admissibility policy.**

| Feature class | Examples | Used by offline selector? | Role |
|---|---|---|---|
| Prefix/format/schema features | JSON node count, max depth, format score, argument-name counts | Yes | Online structural evidence for tool/function/JSON routes |
| Early trajectory features | early logprob, acceptance slope, first/second/third-token deltas | Yes | Online confidence and stability signal |
| Baseline uncertainty features | top-k probability gaps, baseline acceptance statistics | Yes | Online uncertainty and fallback signal |
| Divergence/energy features | first divergence, local divergence ratios, retained energy, keep dimension | Exploratory; timing must be frozen | Candidate/baseline separation and FFN-state diagnostics |
| Post-fork/ post-generation diagnostics | cross-PPL on generated AG tokens, post5 overlap, change-energy fields | No | Explanation only; not used for prospective claims |
| Label-derived or correctness features | answer score, evaluator outcome, post-hoc best candidate | No | Forbidden for deployable claims |

This policy is intentionally conservative. It lets the paper report oracle headroom, offline-selector capture, and diagnostic explanations without collapsing them into the same evidence level. The strongest current selector evidence is therefore not "AG always helps"; it is that exploratory Qwen3.5 offline rankers recover useful candidates on selected tool-structured routes. These are candidate-space results under offline feature evaluation; prospective gates require decision-time feature validation and baseline fallback.

## 4.9 Exploratory grouped random-forest protocol

We evaluate offline RF candidate ranking on Qwen3.5-9B across the six main tasks: 8,886 unique task/sample units, each with twelve old-position AG candidates. Units are assigned to five groups before fitting; all candidates for one unit remain together. The 24-candidate union is used separately for the ConvFinQA selector, not for this RF comparison.

Sorted unit identifiers are shuffled with NumPy RandomState(42) and divided into folds of 1,778, 1,777, 1,777, 1,777, and 1,777 units. Each training fold fits a 200-tree forest with maximum depth 5, minimum leaf size 20, balanced class weights, and random seed 42. Training uses strict fix versus strict harm candidate rows; neutral rows are excluded. The 34-feature contract is fixed, and scaler parameters are fitted only on training-fold rows.

The linear comparator derives feature weights from fix--harm Spearman correlations on each training fold and applies the weighted z-score ranker to held-out units. At test time, both methods score all twelve AG candidates and select the highest-ranked candidate. Candidate rows do not cross folds, but the experiment has no explicit baseline action or abstention rule: it forces one AG choice for every test unit.

The experiment uses saved offline candidate-trajectory features and forced AG choice. We report each fold, the sample-weighted overall effect, and descriptive fold standard deviation. Native and strict effects, and the paired RF-minus-linear contrast, use 10,000 paired within-task bootstrap resamples of held-out units. These intervals condition on the fitted policies and fixed split; they do not include variation from refitting or changing partitions. Acceptance, baseline fallback, feature timing, and serving cost require prospective evaluation.

## 5. Experiment Results: Cross-Model and Cross-Task AG Effects

The decode-only design also shapes interpretation. In the text/tool protocol, AG can alter the trajectory beginning with the second generated token. Single-token choice outputs are mathematically baseline-equivalent under that guard; nonzero historical choice effects belong to prefill-on experiments. The September equivalence controls are reported separately in Section 6.3 and are derived controls, not additional model-generation measurements.

### 5.1 Qwen3.5-9B: tool/function-call is strongest

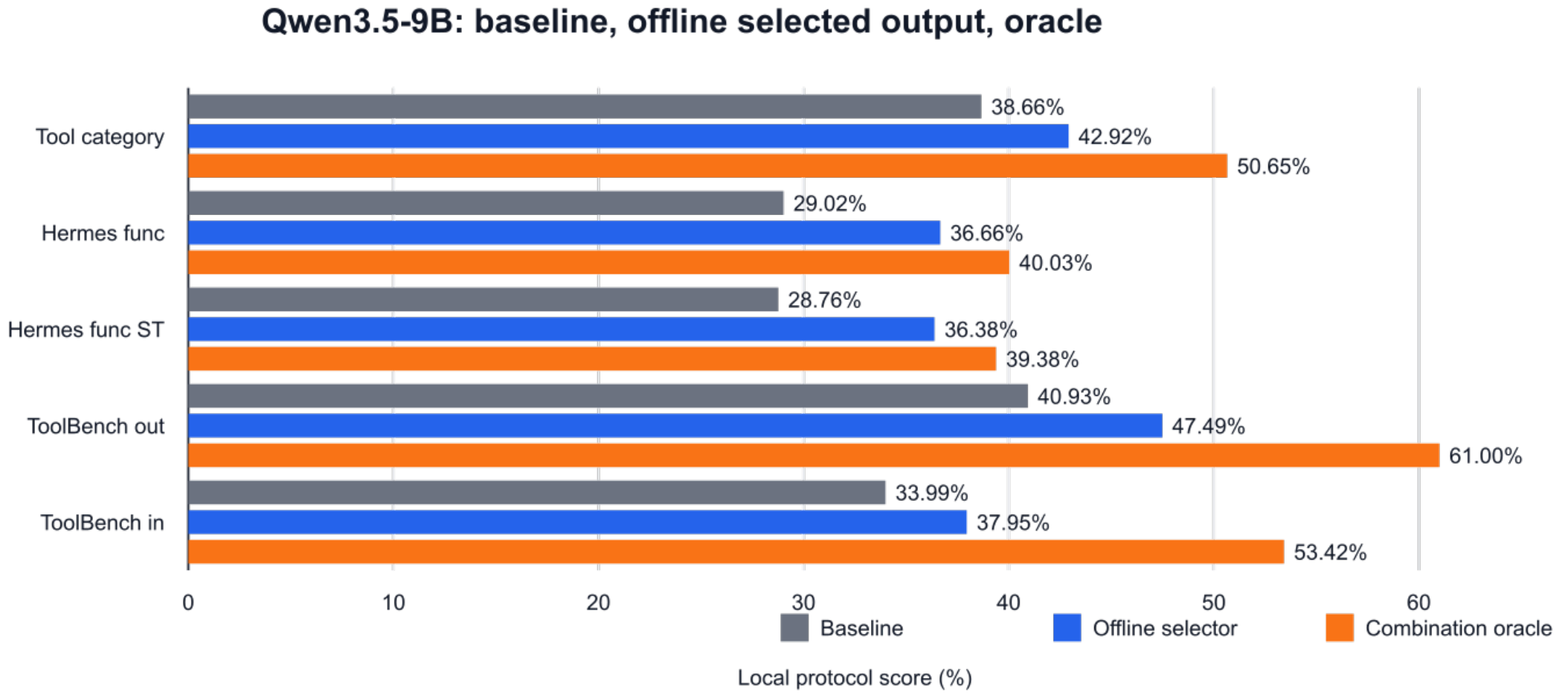

Original Qwen3.5 offline-selector protocol

*Figure 7. Qwen3.5-9B tool-structured results: baseline, exploratory offline selector, and oracle score by route.*

Qwen3.5-9B shows the clearest exploratory selector signal in tool-structured tasks.

**Table 9. Qwen3.5 tool-route classical score: descriptive category micro-average, selector, and oracle headroom.**

| Category/ method | Baseline | Offline selected | Lift | Oracle delta | Oracle capture |
|---|---|---|---|---|---|
| Tool/structured, category mixed selector | 38.66% | 42.92% | +4.27 pp | +11.99 pp | 35.6% |
| Tool/structured, category ridge selector | 38.66% | 42.13% | +3.48 pp | +11.99 pp | 29.0% |

The corresponding combination-oracle upper bound is 50.65%, i.e. +11.99 pp over baseline. It is reported as candidate-space headroom rather than as a deployable method.

The same-output PFHE-format refresh in Section 5.5 gives the corresponding safety view. On the 2,556 selected tool-route units, the Qwen3.5-9B mixed selector has 48 strict fixes, 26 harms, 294 preserved-correct cases, and 2,188 preserved-wrong cases. Its strict accuracy change is +0.86 pp, whereas its native-score change is +4.27 pp. These are complementary endpoints over the same records; the larger native-score gain reflects partial-credit movement that is not a full-credit transition.

At the dataset level, Hermes function-call tasks are the strongest current exploratory offline-selector evidence.

**Table 10. Qwen3.5 dataset-level classical effects for the reported exploratory selectors.**

| Dataset | Baseline | Combination oracle | Offline selected | Selected lift | Oracle capture |
|---|---|---|---|---|---|
| Hermes function call | 29.02% | 40.03% | 36.66% | +7.64 pp | 69.4% |
| Hermes function call single-turn | 28.76% | 39.38% | 36.38% | +7.62 pp | 71.7% |
| ToolBench out-domain | 40.93% | 61.00% | 47.49% | +6.56 pp | 32.7% |
| ToolBench in-domain | 33.99% | 53.42% | 37.95% | +3.96 pp | 20.4% |

These results show that the Qwen3.5-9B candidate matrix contains sample-varying signal. Both learned rows force one AG candidate, using baseline solely as the evaluation reference. They are exploratory offline selectors; prospective gates must additionally evaluate acceptance and baseline fallback.

For financial QA, we refit the offline selector on ConvFinQA alone using 24 candidates per sample. The fixed hash split gives 957 training and 533 test units. Feature ranking, moments, mixed-feature choice, and ridge parameters are estimated on training units only, using the recorded offline feature contract. Table 10a reports the held-out scores. The mixed selector gains +0.75 pp (4 strict fixes, no harms), while ridge gains +0.94 pp (5 fixes, no harms). The exact paired McNemar p-values are 0.125 for mixed selection and 0.0625 for ridge, so neither establishes a strict-accuracy benefit at the 0.05 level. The positive percentile-bootstrap intervals should be read cautiously with so few changed cases. These are small absolute gains from seven baseline-correct test cases and should not be generalized beyond this endpoint.

**Table 10a. ConvFinQA-only offline selector refit on 533 held-out units. Intervals are paired 95% bootstrap intervals conditional on the fixed split and fitted policy.**

| Method | Baseline | Selected | Native lift [95% CI], pp | Strict fix/harm |
|---|---|---|---|---|
| Train-selected fixed | 1.31% | 2.06% | +0.75 [+0.19, +1.50] | 4 / 0 |
| Mixed selector | 1.31% | 2.06% | +0.75 [+0.19, +1.50] | 4 / 0 |
| Ridge selector | 1.31% | 2.25% | +0.94 [+0.19, +1.88] | 5 / 0 |
| Combination oracle (upper bound) | 1.31% | 3.56% | +2.25 [+1.13, +3.56] | 12 / 0 |

The fixed-configuration audit gives a more granular view of where NS AG helps. Table 11 now evaluates the original reported configurations at their corrected physical sites against each September baseline; it does not reselect a winner from the new candidate pool. Consequently the old six/five/tie winner count is not carried forward. C-Eval [10] and MMLU-Pro [9] are prefill-on choice diagnostics; HLE [11] mixes 591 choice and 1,909 generated-answer items. The hand-eight-shot GSM8K supplement is reported under its separately identified prompt protocol.

**Table 11. September scores at the original reported physical configurations (not newly selected best configurations).**

| Dataset | Protocol/category | Baseline | Non-NS fixed | Lift | NS fixed | Lift | Sites non-NS; NS |
|---|---|---|---|---|---|---|---|
| C-Eval | prefill-on choice | 79.2719% | 79.4948% | +0.2229 pp | 79.3462% | +0.0743 pp | P2/e.93; P2/e.93 |
| HLE proxy | prefill-on / mixed | 5.0800% | 5.3200% | +0.2400 pp | 4.9600% | -0.1200 pp | P1v/e.93; P1v/e.93 |
| MMLU-Pro | prefill-on choice | 53.4824% | 53.5073% | +0.0249 pp | 53.5572% | +0.0748 pp | P1v/e.93; P1v/e.93 |
| ConvFinQA dev | Financial QA | 1.2752% | 1.9463% | +0.6711 pp | 2.1477% | +0.8725 pp | P2b/e0.93; P2b/e0.93 |
| Hermes function call | Tool/structured | 28.8795% | 31.4934% | +2.6139 pp | 30.2265% | +1.3470 pp | P1/e0.93; P1/e0.93 |
| Hermes FC single-turn | Tool/structured | 28.8747% | 31.5453% | +2.6707 pp | 30.1430% | +1.2684 pp | P1/e0.93; P1/e0.93 |
| Hermes JSON | Tool/structured | 72.3849% | 72.5107% | +0.1258 pp | 72.4701% | +0.0852 pp | P2b/e0.95; P1/e0.95 |
| ToolBench in-domain | Tool/structured | 34.1940% | 35.8312% | +1.6373 pp | 36.4610% | +2.2670 pp | P1/e0.93; P1/e0.93 |
| ToolBench out-domain | Tool/structured | 43.4059% | 46.2228% | +2.8169 pp | 45.1985% | +1.7926 pp | P2b/e0.93; P1/e0.95 |

All lifts use unrounded scores before display rounding. NS uses w4. P1v is a diagnostic physical control, not shared P1. The original GSM8K Table 11 values (89.39% baseline, 89.54% non-NS, 88.78% NS) are retained as a separate historical protocol summary; the new hand-eight-shot baseline is 87.9454%. These protocol summaries are not treated as matched replications. Fixed configurations shown here do not establish train-selected or held-out optimality.

**Fixed-route effects after allowing for baseline variation.** Table 11a uses the two observed executions of the original fixed configurations from Table 11. Positive margins cannot be attributed solely to choosing the lower of the two observed baseline means. All quantities use unrounded source means; historical shared-input runs are mapped from archival `p1s` to canonical P1. This two-run comparison covers the displayed shared historical/September routes; true gate-only P1s is evaluated separately in the September matrix.

**Table 11a. Qwen3.5 fixed-route sensitivity to observed historical/September variation. Gains and margins are in pp; the margin is the lower AG mean minus the higher baseline mean across the two runs, not a statistical interval. Configurations are those in Table 11.**

| Dataset | N | Non-NS gain historical / September | Non-NS margin | NS gain historical / September | NS margin |
|---|---|---|---|---|---|
| ConvFinQA dev | 1,490 | +0.6040 / +0.6711 | +0.6040 | +0.9396 / +0.8725 | +0.8725 |
| Hermes function call | 1,893 | +2.4297 / +2.6139 | +2.4297 | +1.4059 / +1.3470 | +1.3329 |
| Hermes FC single-turn | 1,893 | +2.6754 / +2.6707 | +2.6533 | +1.4462 / +1.2684 | +1.2510 |
| Hermes JSON | 1,241 | +0.0688 / +0.1258 | +0.0688 | +0.1212 / +0.0852 | +0.0406 |
| ToolBench in-domain | 1,588 | +1.3854 / +1.6373 | +1.2594 | +1.3854 / +2.2670 | +1.3854 |
| ToolBench out-domain | 781 | +3.0730 / +2.8169 | +2.8169 | +2.4328 / +1.7926 | +1.7926 |

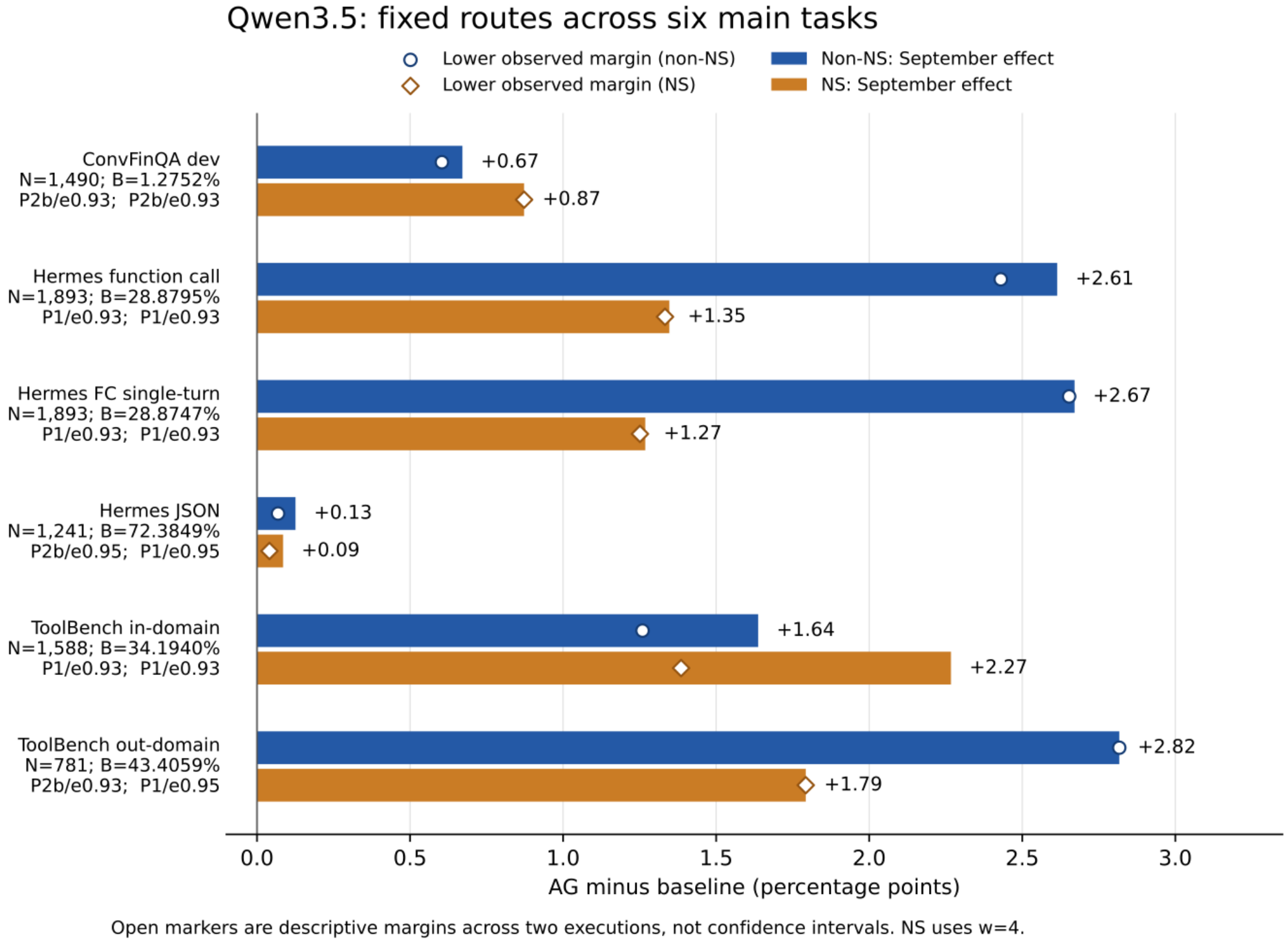

*Figure 8. Current Qwen3.5 fixed-route effects on all six main text/tool tasks. Bars show September AG-minus-baseline local protocol-score changes; open points show L_obs, the lower observed AG mean minus the higher observed baseline mean. Labels give sample counts, September baseline B, and the site/e for each frozen route. NS uses w=4. Margins are descriptive, not confidence intervals; configurations differ by task and family. Exact scores are in Tables 11 and 11a.*

The non-NS function-call margins remain +2.4297 and +2.6533 pp, and the ToolBench margins remain +1.2594 and +2.8169 pp. Thus these fixed-route improvements survive the observed baseline and AG variation. Hermes JSON has only small positive margins. The table supports robustness to the observed executions rather than statistical significance over inputs or future runs; no selector, RF, PFHE, or bootstrap estimate is inferred from these marginal means.

The fixed rows expose a matched P-by-e result inside the refreshed Qwen3.5 tool reports. In the displayed non-NS slice, shared P1/e0.93 gives the strongest function-call score, while genuinely gate-only P1s gives a smaller positive effect. Under the current experimental conditions, P2a and P2b score below baseline on that task. ToolBench out-domain instead favors P2b/e0.93 among the displayed cells, and P2a changes sign across the two e-values. Thus the position-sensitive conclusion is retained but its shared-input versus gate-only attribution is corrected. Tables 20--21 include both sites explicitly.

**Table 12. Refreshed representative Qwen3.5 non-NS slices; descriptive comparisons within the displayed grid.**

| Task | Displayed leading cell | Same-site e0.95 | Contrast |
|---|---|---|---|
| Hermes function call | P1/e0.93: 31.4934% (+2.6139 pp) | P1/e0.95: 30.2473% (+1.3678 pp) | True P1s/e.93: 29.3844% (+0.5048 pp) |
| ToolBench out-domain | P2b/e0.93: 46.2228% (+2.8169 pp) | P2b/e0.95: 44.8143% (+1.4085 pp) | Shared P1 and gate-only P1s are positive; P2a changes sign |

## 5.2 Qwen3-8B: source-separated headroom without a fixed tool route

The archived Qwen3 protocols are analyzed separately, with source-specific sample domains, baseline scores, and correctness outcomes. Fixed-route and oracle results are interpreted within each domain; learned-selector and Hermes-JSON conclusions are not drawn from a pooled candidate matrix.

**Table 13. Qwen3 source-separated classical and PFHE-format record-level results on the tool category. Native values are descriptive heterogeneous-scorer micro-averages.**

**(a) Native task effects.**

| Qwen3 protocol | Method | Test N | Native B to I | Native delta (95% paired CI) |
|---|---|---|---|---|
| Old-P, full | Fixed P2/e0.95 selected on train | 2,556 | 42.76% to 42.65% | -0.11 pp [-0.60, 0.38] |
| Old-P, full | 12-candidate oracle | 2,556 | 42.76% to 49.63% | +6.87 pp [6.15, 7.60] |
| New-P, restricted subset | Fixed NS-P2b/ e0.93/w4 selected on train | 855 | 49.41% to 49.01% | -0.41 pp [-1.81, 1.03] |
| New-P, restricted subset | 12-candidate oracle | 855 | 49.41% to 58.29% | +8.88 pp [7.40, 10.50] |

**Table 13(b). Strict full-credit transitions on the same units as panel (a). N and configuration identities are unchanged.**

| Protocol / method | Fix A | Harm H | Preserved correct C | Preserved wrong D | Strict delta (95% paired CI) |
|---|---|---|---|---|---|
| Old-P / fixed | 10 | 15 | 364 | 2,167 | -0.20 pp [-0.59, 0.20] |
| Old-P / oracle | 60 | 0 | 379 | 2,117 | +2.35 pp [1.76, 2.93] |
| New-P / fixed | 16 | 6 | 142 | 691 | +1.17 pp [0.12, 2.22] |
| New-P / oracle | 37 | 0 | 148 | 670 | +4.33 pp [3.04, 5.73] |

The native category values are descriptive micro-averages over five heterogeneous scorers; source-specific dataset endpoints remain primary. Table 13 is an exploratory PFHE-format analysis of paired records within the frozen matrices, interpreted under each source's scoring protocol.

Both sources contain oracle headroom, but neither train-selected fixed route has a positive native-score interval at the tool-category level. The new-P fixed row is also a useful PFHE-format counterexample: 16 strict fixes versus 6 harms produce a positive strict-accuracy effect, while the native-score mean is negative. Classical magnitude and strict transitions therefore cannot be collapsed into one success label.

The new-P rows use only 855 tool test units and have a much higher baseline than the full old-P test domain. They are restricted-subset evidence, not a matched ablation against old-P. Within the full old-P protocol, ToolBench in-domain and out-domain have positive fixed native effects (+1.98 pp and +2.70 pp), whereas Hermes JSON is only +0.96 pp with an interval crossing zero. These rows support localized fixed effects and candidate headroom, not the excluded pooled learned headline.

### 5.3 Qwen2.5-7B: source-specific oracle headroom, near-zero fixed effects

The Qwen2.5 old-P and new-P sources have different baseline scores and correctness outcomes. Pooling them would attach interventions to inconsistent reference outcomes. We instead analyze the two 12-candidate sources independently.

**Table 14. Qwen2.5 source-separated classical and PFHE-format record-level results on the tool category. Native values are descriptive heterogeneous-scorer micro-averages.**

**(a) Native task effects.**

| Qwen2.5 protocol | Method | Test N | Native B to I | Native delta (95% paired CI) |
|---|---|---|---|---|
| Old-P, full | Fixed P1v/e0.95 selected on train (raw p1) | 2,556 | 44.18% to 43.96% | -0.22 pp [-0.48, 0.04] |
| Old-P, full | 12-candidate oracle | 2,556 | 44.18% to 47.10% | +2.92 pp [2.53, 3.34] |
| New-P, full | Fixed NS-P1/e0.93/w4 selected on train (raw p1s) | 2,556 | 44.16% to 44.11% | -0.05 pp [-0.41, 0.33] |
| New-P, full | 12-candidate oracle | 2,556 | 44.16% to 47.74% | +3.59 pp [3.16, 4.06] |

**Table 14(b). Strict full-credit transitions on the same units as panel (a). N and configuration identities are unchanged.**

| Protocol / method | Fix A | Harm H | Preserved correct C | Preserved wrong D | Strict delta (95% paired CI) |
|---|---|---|---|---|---|
| Old-P / fixed | 2 | 4 | 337 | 2,213 | -0.08 pp [-0.27, 0.12] |
| Old-P / oracle | 21 | 1 | 340 | 2,194 | +0.78 pp [0.43, 1.13] |
| New-P / fixed | 14 | 4 | 335 | 2,203 | +0.39 pp [0.08, 0.70] |
| New-P / oracle | 31 | 0 | 339 | 2,186 | +1.21 pp [0.82, 1.64] |

As above, these native category values are descriptive heterogeneous-scorer micro-averages. Table 14 is an exploratory PFHE-format analysis of paired records within the frozen matrices under source-specific scoring protocols.

Both sources show candidate-space headroom, but neither train-selected fixed route has a positive native-score interval. The new-P fixed row again shows why the PFHE-format view is complementary: its 14/4 strict fix--harm balance is positive, yet the mean native effect is -0.05 pp with an interval spanning zero. On ToolBench out-domain, both independently selected source protocols give a borderline +1.16 pp fixed effect; that localized row does not establish a positive category policy or reconcile the incompatible baselines.

The defensible Qwen2.5 conclusion is therefore narrower than a pooled learned- gate failure. Useful alternatives exist in each source-specific candidate space, but a fixed route selected on training data does not provide reliable tool-category improvement. Extending these comparisons to learned selection requires a shared baseline protocol and candidate universe.

## 5.4 NS AG versus non-NS AG

The September full-domain summary below uses sample-count-weighted native means of the per-dataset configurations frozen in Table 11. It is descriptive and mixes heterogeneous native scorers; it is neither a single configuration applied to an entire category nor a learned selector. Tool/structured favors non-NS; the financial-QA endpoint favors NS.

**Table 15. September descriptive full-domain summary of the frozen per-dataset Table 11 routes.**

| Category | N | Baseline | Non-NS | Lift | NS | Lift | Higher mean |
|---|---|---|---|---|---|---|---|
| Tool/structured | 7396 | 38.8532% | 40.8759% | +2.0227 pp | 40.2129% | +1.3597 pp | non-NS |
| Financial QA (ConvFinQA) | 1490 | 1.2752% | 1.9463% | +0.6711 pp | 2.1477% | +0.8725 pp | NS |

Table 15 spans 8,886 evaluation units across the six main tasks. Each category weights the frozen Table 11 routes by sample count. This fixed-route summary is separate from the held-out ConvFinQA selector and the 2,556-unit tool selector. The Qwen3 and Qwen2.5 findings retain the source-separated interpretations in Sections 5.2 and 5.3.

### 5.5 PFHE-format record-level results: paired benefit--risk behind the classical scores

We calculated the companion paired view directly from the selected-output records used by the exploratory Qwen3.5 category rows. The analysis covers 2,556 tool-route units with unique, finite baseline/selected-output records. The table reports the four-state PFHE-format vector together with its linked strict and native effects. Intervals use 10,000 paired bootstrap replicates with deterministic seeds derived by SHA-256 from each result-group key. These are exploratory record-level score comparisons under the stated historical protocol; prospective runtime endpoints form a separate evaluation.

**Table 16. Qwen3.5 strict PFHE-format transitions on 2,556 selected-output records per selector. Rates are counts divided by N; native effects for the same outputs follow below.**

| Selector | Fix A | Harm H | Preserved correct C | Preserved wrong D | Strict delta (95% paired CI) |
|---|---|---|---|---|---|
| Mixed | 48 | 26 | 294 | 2,188 | +0.86 pp [0.20, 1.53] |
| Ridge | 64 | 40 | 280 | 2,172 | +0.94 pp [0.16, 1.72] |

For the same selected outputs, native gains are +4.27 pp [3.51, 5.02] for mixed selection and +3.48 pp [2.57, 4.40] for ridge (paired 95% intervals).

The identities in Section 4.7 audit every row. For Qwen3.5-9B, for example, baseline strict accuracy is `(26+294)/2556=12.52%`, selected-output strict accuracy is `(48+294)/2556=13.38%`, and their +0.86 pp difference equals `(48−26)/2556`. The positive native effect is larger because many function-call examples improve without reaching exact full credit. Qwen3-8B illustrates the same native-versus-strict distinction under the source-separated fixed result in Section 5.2, but its excluded pooled learned output is not used here.

Derived PFHE-format rates make the operational trade-off explicit.

**Table 17. Derived PFHE-format transition, recovery, regression-risk, and changed-case rates.**

| Learned selector | Transition coverage `(A+H)/N` | Baseline-error recovery `A/(A+D)` | Regression risk `H/(H+C)` | Fix share `A/(A+H)` |
|---|---|---|---|---|
| Qwen3.5-9B category mixed | 2.90% | 2.15% | 8.13% | 64.86% |
| Qwen3.5-9B category ridge | 4.07% | 2.86% | 12.50% | 61.54% |

Strict transitions are intentionally not used as a substitute for native partial-credit magnitude. The mixed selector has 684 positive, 287 negative, and 1,585 tied native-score differences; ridge has 720/346/1,490. The paired median is zero for both because unchanged scores are the majority. The native gains therefore arise from a minority of changed cases, and the higher ridge transition coverage also brings a higher regression risk.

The dataset-specific magnitude view below uses each Qwen3.5 dataset's reported learned method from Section 5.1. It is not pooled with the category-selector PFHE-format table and does not imply that one policy generated every row.

**Table 18. Qwen3.5 dataset-level paired native-score magnitude and direction counts.**

| Model / tool dataset | N | Native delta (95% paired CI) | Positive / negative / tied |
|---|---|---|---|
| Qwen3.5 / Hermes function call | 649 | +7.64 pp [6.61, 8.70] | 319 / 94 / 236 |
| Qwen3.5 / Hermes function call single-turn | 662 | +7.62 pp [6.59, 8.66] | 337 / 94 / 231 |
| Qwen3.5 / Hermes JSON mode | 430 | +0.26 pp [-0.08, 0.62] | 31 / 18 / 381 |
| Qwen3.5 / ToolBench in-domain | 556 | +3.96 pp [1.08, 6.83] | 45 / 23 / 488 |
| Qwen3.5 / ToolBench out-domain | 259 | +6.56 pp [2.32, 11.20] | 26 / 9 / 224 |

These paired magnitude rows preserve the classical conclusion rather than replace it: Qwen3.5-9B is strongest on function calling, while the source- separated Qwen3/Qwen2.5 fixed results remain cross-model boundaries. The PFHE-format view adds the fact that effect magnitude, transition coverage, and regression risk are distinct properties. Historical `partial10` totals are excluded because the source artifacts use incompatible raw-delta and decile-crossing definitions.

## 5.6 Exploratory random-forest candidate selection

Table 19 reports the six-task grouped five-fold RF comparison. Native delta is the selected candidate score minus its paired baseline score. Task-native scorers differ, so the pooled mean is a descriptive micro-average; per-task results are saved with the paired records.

**Table 19. Six-task grouped five-fold RF comparison on Qwen3.5. Overall effects weight units; +/- reports fold SD (ddof=0), not a confidence interval.**

| Fold | Evaluation units | Linear native delta | Linear strict fix/harm | Base RF native delta | Base RF strict fix/harm |
|---|---|---|---|---|---|
| 0 | 1,778 | -0.49 pp | 19 / 14 | -0.02 pp | 21 / 11 |
| 1 | 1,777 | +0.17 pp | 30 / 12 | +0.72 pp | 34 / 14 |
| 2 | 1,777 | -0.02 pp | 32 / 14 | +0.46 pp | 35 / 12 |
| 3 | 1,777 | +0.56 pp | 32 / 8 | +1.21 pp | 40 / 5 |
| 4 | 1,777 | -0.75 pp | 16 / 18 | -0.06 pp | 23 / 16 |
| Overall effect +/- fold SD; total | 8,886 | **-0.11 +/- 0.47 pp** | **129 / 66** | **+0.46 +/- 0.48 pp** | **153 / 58** |

Under the current experimental conditions, RF has positive native deltas in three of five folds. Across all 8,886 held-out units, baseline native score is 32.61%, RF-selected score is 33.07%, and the paired delta is +0.46 pp (95% CI [0.13, 0.80]). The linear comparator has -0.11 pp [-0.43, 0.22]. The paired RF-minus-linear contrast is +0.57 pp [0.31, 0.82]. Thus this fixed-split comparison supports a positive conditional native-score effect and a positive RF--linear contrast; it does not establish stability across partitions, new datasets, or checkpoints.

The RF-selected records directly give 153 fixes, 58 harms, 923 preserved-correct cases, and 7,752 preserved-wrong cases. Baseline strict accuracy is 11.04%, RF-selected strict accuracy is 12.11%, and the paired change is +1.07 pp [0.75, 1.38]. The strict RF-minus-linear contrast is +0.36 pp [0.12, 0.60]. These record-level transitions accompany the native effect; strict correctness and partial-credit magnitude remain different endpoints.

The intervals above are conditional on one grouped partition and the fitted policies. They quantify held-out unit variation, not variation in training data, random seeds, or future task mixtures. RF remains an exploratory offline selector: the current experiment does not validate neutral-outcome modeling, an acceptance decision, baseline fallback, cross-model transfer, or deployment. Repeated grouped partitions and independent evaluations are needed to assess the stability of the observed advantage.

# 6. Analysis: Mechanisms, Boundaries, and Gate Limits

## 6.1 Position and e-value sensitivity

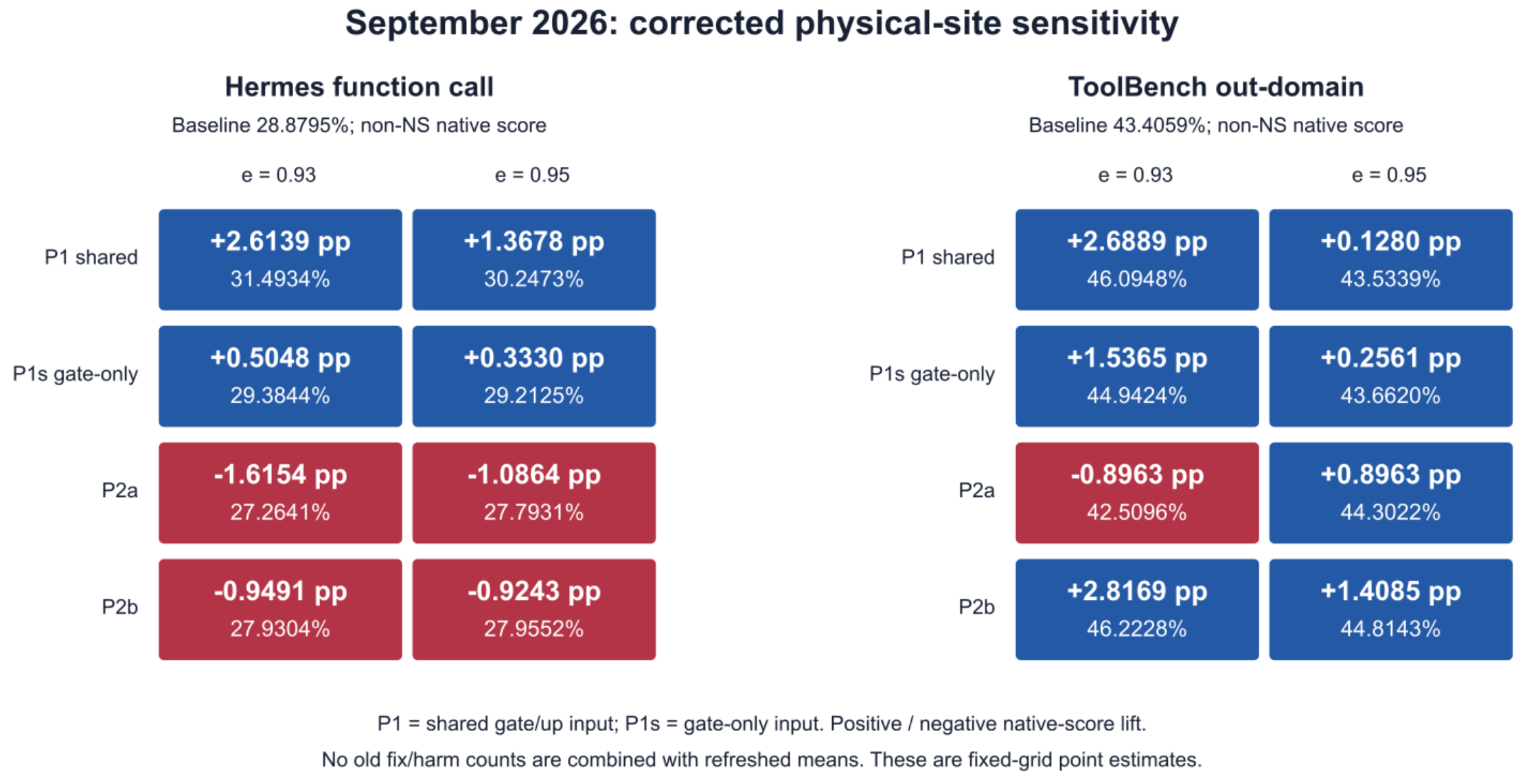

*Figure 9. Refreshed Qwen3.5 non-NS position-by-e sensitivity. Shared P1 and true gate-only P1s are separate rows; all cells use current native scores and the current task baseline. No historical fix/harm counts are attached to new means.*

Under the current experimental conditions, AG behavior depends on both position and e. The September rerun confirms a qualitative distinction between shared-input P1 and gate-only P1s. The large function-call gain formerly attributed to P1s belongs to shared P1. Gate-only P1s remains positive in these non-NS cells but is appreciably weaker; its NS function-call cells are negative. Within these tested conditions, the leading site depends on the task; this does not establish a general limitation of P2a or other sites.

**Table 20. September Qwen3.5 Hermes function call non-NS position-by-e sensitivity; baseline 28.8795%.**

| Position | e=0.93 candidate | e=0.95 candidate | Local readout |
|---|---|---|---|
| P1 shared | 31.4934% (+2.6139 pp) | 30.2473% (+1.3678 pp) | Largest displayed function-call gain |
| P1s gate-only | 29.3844% (+0.5048 pp) | 29.2125% (+0.3330 pp) | Positive but smaller |
| P2a | 27.2641% (-1.6154 pp) | 27.7931% (-1.0864 pp) | Below baseline here |
| P2b | 27.9304% (-0.9491 pp) | 27.9552% (-0.9243 pp) | Below baseline here |

ToolBench out-domain gives a different profile: P2b/e0.93 remains the strongest displayed cell, shared P1 and gate-only P1s remain positive, and P2a changes sign. This is a descriptive fixed-grid comparison, not held-out selection of a universally optimal site.

**Table 21. September Qwen3.5 ToolBench out-domain non-NS position-by-e sensitivity; baseline 43.4059%.**

| Position | e=0.93 candidate | e=0.95 candidate | Local readout |
|---|---|---|---|
| P1 shared | 46.0948% (+2.6889 pp) | 43.5339% (+0.1280 pp) | Positive at both e values |
| P1s gate-only | 44.9424% (+1.5365 pp) | 43.6620% (+0.2561 pp) | Positive at both e values |
| P2a | 42.5096% (-0.8963 pp) | 44.3022% (+0.8963 pp) | Sign changes with e |
| P2b | 46.2228% (+2.8169 pp) | 44.8143% (+1.4085 pp) | Highest displayed e0.93 score |

**Why might value-based P2b help on ToolBench?** In the SwiGLU update (Section 3.5), `f_j=g_j u_j`, with `g=SiLU(W_g x)` and `u=W_u x` [2]. P2b selects a diagonal mask `D_u` from u, giving `y'=W_d D_u f`. Its selection signal resolves expanded value channels before multiplication by g; the gate still controls the contribution of every retained channel. This may suppress distracting value channels while preserving gate-dependent modulation.

The alternatives expose different signals. Earlier P1 masking precedes both projections: changing one input coordinate can perturb many gate and value channels. P1s confines that input change to the gate branch. P2a ranks the post-SiLU gate signal, while later P2 ranks the fused product f, where gate and value contributions are already entangled. P3 acts after `W_d` has mixed expanded channels into output coordinates. P2b may therefore offer a useful balance between channel resolution and preservation of gate modulation. This is a hypothesis about adaptive mask selection, not evidence that earlier or later intervention is generally inferior.

Section 3.6 gives the key constraint: an identical fixed mask on g, u, or their product is algebraically equivalent. Any P2a/P2b/P2 difference must therefore arise from how their signals and histories select masks, including downstream autoregressive effects, rather than multiplication order alone. Hermes provides a counterexample to a universal P2b advantage (Table 20). Controlled mask-transfer and matched-sparsity ablations are needed to test the proposed explanation.

September native means and archived strict fix/harm counts describe different protocols and are not paired. The old-P/new-P cross-model restrictions apply to the archived analyses. Position, retained mass, algorithm variant, model, and task jointly determine whether an AG candidate helps.

Genuine gate-only P1s is not ineffective in general. On ToolBench in-domain, NS/e0.93 reaches 36.6499% (+2.4559 pp); on ToolBench out-domain, NS/e0.95 reaches 45.0704% (+1.6645 pp). These are new fixed-configuration point estimates, not historical P1s replications or selected-policy guarantees.

**Baseline variation changes the strength of the conclusion.** Greedy generation did not make the baseline invariant. The two-run margins in Table 11a use the weaker observed AG score and stronger observed baseline score. The positive function-call and ToolBench margins show that their observed gains persist under this conservative comparison; JSON gains remain small. These margins describe robustness to the observed executions, while formal significance over future runs requires a repeated-run design. Historical selector confidence intervals cannot be attached to the September fixed-route means.

## 6.2 Interpreting NS versus non-NS AG

The Qwen3.5 fixed-report comparison rejects a universal family ordering. Tool/structured favors non-NS, whereas the ConvFinQA financial-QA endpoint favors windowed NS. This is an intra-protocol descriptive result, not a guarantee that either family transfers to another task or checkpoint.

Mechanistically, non-NS AG ranks dimensions from the current activation or entropy signal. It may preserve a sharp local format decision, such as selecting a function-call slot. NS AG changes only that ranking score: with `w4`, recent activation states form a short local matrix before the same mask interface is applied. This may stabilize a short trajectory at the cost of blurring a useful current-token peak. These explanations are hypotheses consistent with the Qwen3.5 task interaction, not causal conclusions.

The archived Qwen3 and Qwen2.5 sources cannot test cross-model family ordering because their old-P and new-P protocols are not jointly aligned. Their prior NS/non-NS comparison is therefore excluded. A valid extension must select each family within the training portion of one frozen source protocol, evaluate on the identical held-out domain, and report classical and PFHE endpoints together. Until then, retaining both candidate families is a design choice motivated by Qwen3.5 diversity and source-separated oracle headroom, not evidence for a universal model-aware router.

## 6.3 Negative diagnostics and claim boundary

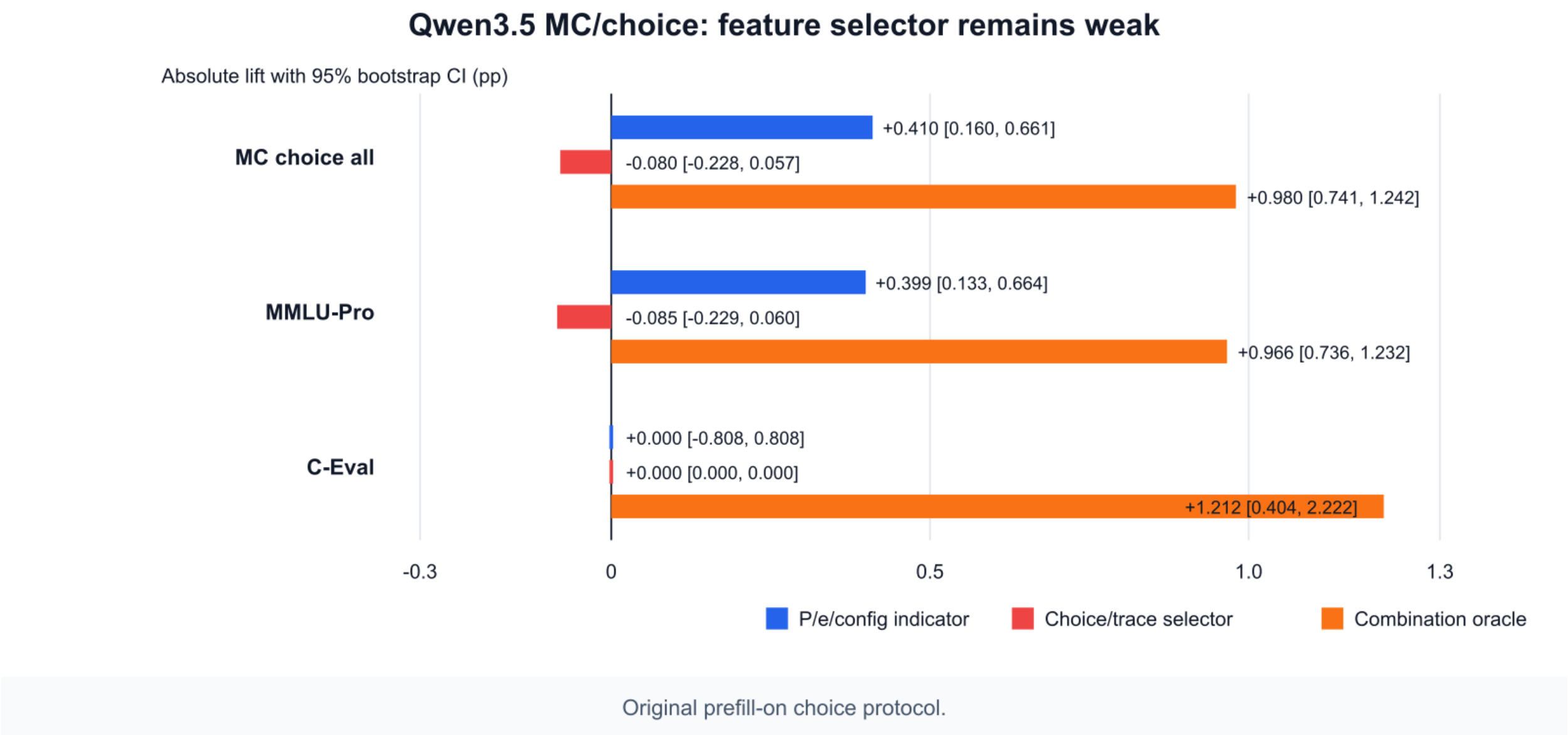

*Figure 10. Historical prefill-on MC/choice diagnostics; the new decode-only identity controls are separate. MC/choice diagnostics define the claim boundary: static P/e signals are small, and the current choice/trace feature gate is not positive.*

The MC track is useful because it constrains the claim. The retained historical prefill-on aligned MC/choice rerun covers C-Eval and MMLU-Pro, built from raw detail JSONs with baseline/candidate alignment and the same stable split policy used by the aligned text/tool reruns. We do not use the older first-batch reports as the source for this conclusion. The rerun slightly narrows the old negative boundary: static P/e/config signals can be positive, yet the current choice/trace feature ranker is not a deployable positive gate.

**Table 22. Qwen3.5 MC/choice classical effects and negative selector boundary.**

| Scope | Baseline | P/e/config indicator | Choice/trace feature selector | Combination oracle | Interpretation |
|---|---|---|---|---|---|
| MC choice all | 54.27% | +0.410 pp, CI [0.160, 0.661] | -0.080 pp, CI [-0.228, 0.057] | +0.980 pp, CI [0.741, 1.242] | Small static signal; feature selector not established |
| MMLU-Pro | 52.85% | +0.399 pp, CI [0.133, 0.664] | -0.085 pp, CI [-0.229, 0.060] | +0.966 pp, CI [0.736, 1.232] | Oracle headroom remains small |
| C-Eval | 77.98% | +0.000 pp, CI [-0.808, 0.808] | +0.000 pp, CI [0.000, 0.000] | +1.212 pp, CI [0.404, 2.222] | High baseline; no learned choice trace signal |

The historical Table 22 coefficients, intervals and selector effects belong to prefill-on choice measurements; they are not evidence of a nonzero decode-only single-token effect. Under the verified decode-only guard, the following new controls are exactly baseline-equivalent.

**Table 22a. September single-token decode-only equivalence controls (mathematical propagation, not new full-dataset model inference).**

| Scope | N | Baseline | Decode-only candidate | Delta |
|---|---|---|---|---|
| C-Eval | 1,346 | 79.2719% | 79.2719% | 0.0000 pp |
| MMLU-Pro | 12,032 | 53.4824% | 53.4824% | 0.0000 pp |
| HLE choice subset only | 591 | 21.1506% | 21.1506% | 0.0000 pp |

These controls cover seven physical sites and two algorithm families, giving 42 configurations and 195,566 explicitly derived records. The 1,909 generated-answer HLE items are excluded. Frozen-hook checks and independent record validation establish the identity under the stated guard, not a full-dataset bitwise inference rerun. Thus multi-token structured generation remains the plausible decode-only target, while prefill-on intervention is a different experiment.

### 6.4 Interpreting structured-output gains

The structured-output gains suggest two hypotheses. Constrained outputs may make small activation changes visible as different function names, argument tokens, or boundary continuations. Their prefixes may also supply useful early-generation features for a future controller. Neither explanation is established by the aggregate scores or validates an online gate.

The local endpoint scores measure lexical overlap or answer/action-name containment. A score gain can therefore reflect a repaired name or string boundary without proving schema validity, successful execution, or end-to-end task completion. Long-form QA imposes different requirements, and its displayed effects are smaller or less stable.

The sample-level audit makes this concrete. Table 23 shows three representative historical ToolBench out-domain examples from the Qwen3.5-9B non-NS P1v/e0.93 (raw p1) run, using first-action exact match rather than the older loose `contains` proxy. The positive examples are not semantic rewrites of the answer; they are format and action-selection repairs. The negative example shows the boundary: AG can also add a spurious chat suffix to an otherwise correct tool name.

**Table 23. Representative ToolBench first-action fixes and harms under a fixed Qwen3.5 AG route.**

| Case | Task cue | Gold first action | Baseline first action | AG first action | Interpretation |
|---|---|---|---|---|---|
| Fix | quote categories | `list_categories_for_famous_quotes` | `list_categories_for_famous_quotesuser` | `list_categories_for_famous_quotes` | AG removed a prompt-boundary suffix that invalidated the action |
| Fix | streaming availability | `search_basic_free_for_streaming_availability` | `search_pro_for_streaming_availability` | `search_basic_free_for_streaming_availability` | AG shifted the selected tool from the pro endpoint to the basic/free endpoint required by the state |
| Harm | streaming availability | `search_pro_for_streaming_availability` | `search_pro_for_streaming_availability` | `search_pro_for_streaming_availabilityuser` | AG polluted a correct action with a boundary suffix, turning a clean hit into a failure |

These cases also explain why the paper avoids relying on loose string-contains metrics for deployment claims. Some raw generations echo parts of the prompt or tool list; first-action parsing and sample-level fix/harm accounting are therefore necessary to distinguish real action repair from evaluator artifacts.

The retained historical aligned Qwen3.5 first-divergence analysis supports a trajectory-fork interpretation of these cases. On its tool-route test split, `first_divergence_step` and `first_divergence_norm` have about 4.81 pp and 4.56 pp bin-delta spread. These are diagnostic associations, not a prospective gate result. Previous cross-model divergence comparisons are omitted because the corresponding Qwen3/Qwen2.5 candidate unions do not pass the protocol alignment gate.

The corresponding historical layer-trace outcome aggregation is more cautious. The September source audit additionally resolves a one-step indexing issue (decode trace 0 corresponds to generated token 1 in zero-based indexing); corrected slices require a continuous valid trace. Archived AG-labelled full-prefix post-fork probabilities bypass decode AG and are excluded as actual AG-trajectory features. Baseline cross-PPL remains an offline diagnostic, not a decision-time-legal feature. On the current Qwen3.5 choice trace, fix and harm rows have very similar layer-trace means: rank-ratio 0.7042 versus 0.7015, residual magnitude 0.0734 versus 0.0727, and entropy 8.3561 versus 8.3982. These values make layer traces useful as mechanism diagnostics, not as a standalone gate. They also prevent the paper from claiming a causal low/mid/high-layer optimum or a universal early-token rule. The safest mechanism statement is therefore: AG can fork the generation trajectory at a discrete decision point, autoregressive decoding can propagate that fork, and current layer-level aggregates are not yet strong enough to serve as independent causal evidence.

### 6.5 Cross-model routing requires protocol alignment first

The cross-model comparison starts with protocol identity: candidate sources must share the same sample universe, baseline, scorer, and split before pooling. Historical Qwen3 old-P and new-P use distinct sample domains and baseline outcomes; historical Qwen2.5 sources also differ in baseline outcomes. Their fixed-route and oracle results are therefore interpreted separately. The September collection supplies its own corrected-site comparisons. These source-specific results do not establish cross-model selector transfer.

A defensible cross-model workflow is therefore: freeze one protocol per model; namespace and group samples before splitting; train and evaluate a selector only inside that protocol; validate an acceptance action against the baseline; and then compare models using explicitly aligned estimands. The retained Qwen3.5 analysis establishes exploratory offline candidate ranking only; it does not complete acceptance/fallback validation. No September cross-model selector or universal AG switch is claimed.

### 6.6 Oracle headroom versus selected or fixed gain

Combination oracle is useful because it measures whether the AG candidate space contains helpful alternatives. It is not deployable because it chooses the right candidate after observing the answer score. A large oracle gap with a weak selected or fixed result means the research problem has not been solved; it does not mean the method is ready.

The source-separated tool-category comparison quantifies the gap.

**Table 24. Non-oracle results versus combination-oracle headroom under aligned or source-separated protocols.**

| Model protocol | Non-oracle result | Combination-oracle headroom | Interpretation |
|---|---|---|---|
| Qwen3.5 aligned 24-candidate union | Mixed selector +4.27 pp | +11.99 pp | Exploratory selector captures part of the aligned headroom |
| Qwen3 old-P, full | Train-selected fixed -0.11 pp | +6.87 pp | Headroom without a positive fixed route |
| Qwen3 new-P, restricted N=855 | Train-selected fixed -0.41 pp | +8.88 pp | Restricted domain; not comparable with old-P |
| Qwen2.5 old-P, full | Train-selected fixed -0.22 pp | +2.92 pp | Headroom without a positive fixed route |
| Qwen2.5 new-P, full | Train-selected fixed -0.05 pp | +3.59 pp | Separate baseline protocol |

The Qwen3.5 selector also forces an AG choice, so even its positive row is not a deployable capture estimate. Table 24's role is to separate useful candidate space from selection and acceptance, not to rank models by oracle score.

### 6.7 Feature families

The current strongest feature families are:

- Prefix/format/schema features, especially for JSON and function-call tasks.
- Early logprob and acceptance trajectory features, such as slope, trend, and

first/second/third token changes.

- Divergence and first-divergence features, useful but not yet sufficient for

long QA.

- Diagnostic post-fork features such as cross-PPL, which explain candidate

plausibility but require care before online deployment.

The Qwen3.5 feature-family rerun uses the aligned meta CSV pool, label/diagnostic exclusions, and the stable hash split. On the 2,556-unit tool test scope, its offline selector variants are CI-positive in native score.

**Table 25. Qwen3.5 exploratory feature-family native-score effects on the tool test scope.**

| Model | P/e/config only | P/e + nonconfidence features | P/e + confidence features | Combination oracle |
|---|---|---|---|---|
| Qwen3.5-9B | +1.756 pp, CI [1.101, 2.421] | +4.555 pp, CI [3.746, 5.470] | +2.989 pp, CI [2.107, 3.866] | +11.994 pp, CI [11.123, 13.002] |

At dataset level, Qwen3.5 Hermes function-call is strong under P/e plus nonconfidence features (+7.514 pp, CI [6.580, 8.525]), while confidence-style baseline uncertainty features are positive but smaller. The nonconfidence group combines prefix/format with other online candidates, so it is not a pure prefix-only ablation. Several features summarize baseline/candidate trajectories and are evaluated offline; prospective use requires measured feature timing and serving cost. The table supports exploratory feature signal within Qwen3.5, not a prospective or cross-model gate.

## 6.8 Gate trade-off curves

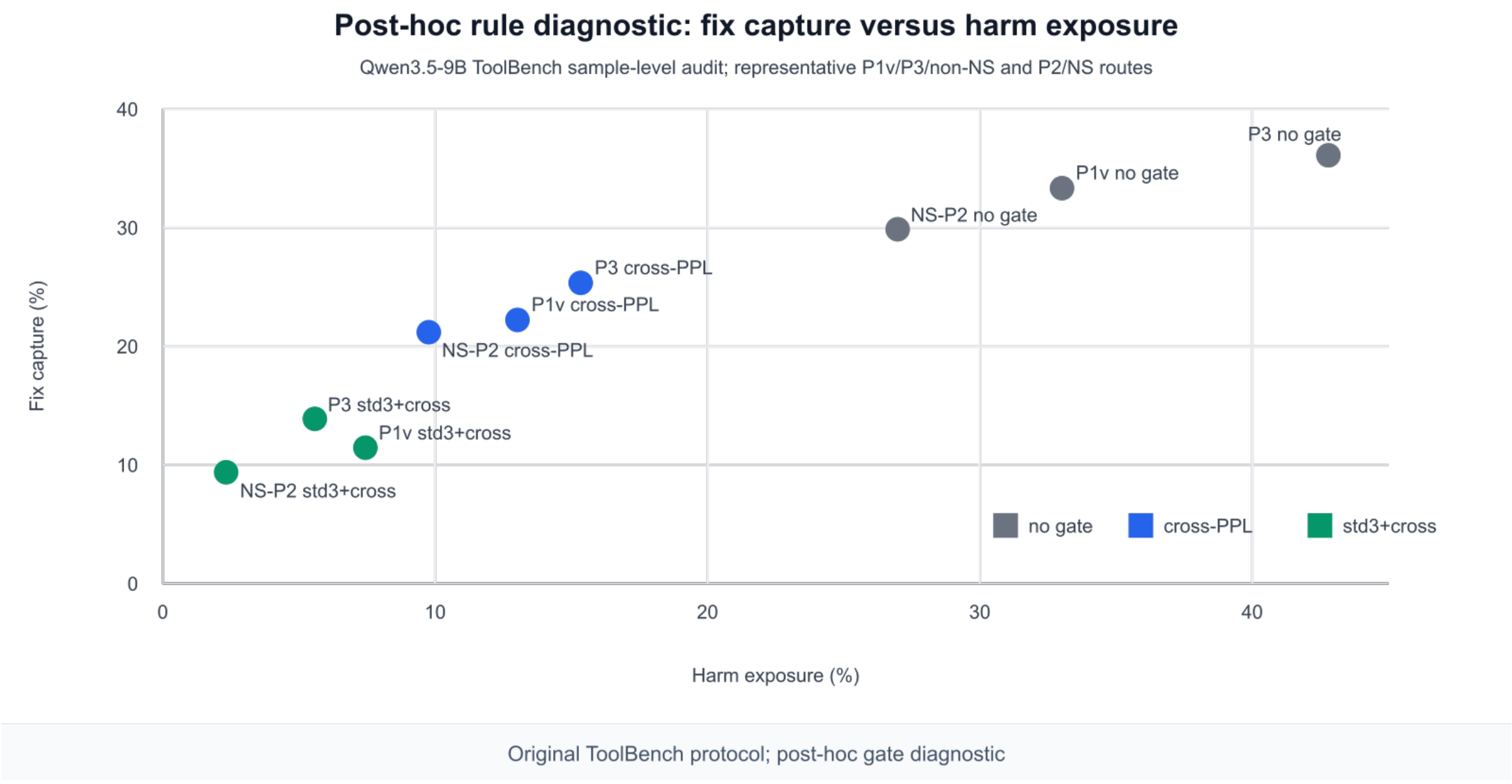

*Figure 11. Gate trade-off diagnostic: stricter post-hoc rules reduce harm exposure while also giving up fix capture. The archived input route is canonical P1v (raw p1), not shared P1.*

Gate choice is a Pareto trade-off rather than a binary question of whether AG helps. The Qwen3.5-9B ToolBench sample-level audit illustrates the pattern. For the historical non-NS P1v/e0.93 route (raw p1), no rule captures 33.3% of available fixes but exposes 33.0% of harms; adding `crossppl <=`

`1.25833` reduces harm exposure to 13.0% while retaining 22.2% fix capture; adding the stricter `std3&crossppl` gate reduces harm exposure further to 7.4% but keeps only 11.5% fix capture. The same qualitative curve appears for non-NS P3/e0.93 and NS P2/e0.93/w4. Because cross-PPL is computed from completed candidate tokens, these operating points are diagnostic and cannot be presented as prospective coverage--risk performance. They motivate, but do not validate, the safety-first, balanced, and performance-first profiles discussed below.

### 6.9 What the RF diagnostic establishes and does not

Under the current experimental conditions, RF is compared with a correlation-weighted linear ranker. Its +0.46 pp native effect and +0.57 pp paired RF--linear contrast have positive conditional intervals in the six-task comparison. Two folds have negative RF native deltas in this partition. The evidence supports a positive within-partition result; generalization to other conditions requires evaluation. These observations do not establish a general limitation of RF.

Three requirements distinguish prospective AG control from this diagnostic. First, training should represent neutral candidate outcomes and explicit baseline utility alongside strict fixes and harms. Second, acceptance should determine whether the best-ranked AG candidate improves on baseline, with abstention when it does not; the present experiment forces an AG choice. Third, trajectory features require a measured decision-time and serving-cost contract before online use.

Fold standard deviation describes variation among the five held-out groups. The paired bootstrap intervals in Section 5.6 condition on their fitted policies; neither statistic represents uncertainty across new training partitions. Repeated grouped splits, independent datasets, cross-model replication, and properly nested stacking are needed before claiming stable transfer or a deployment-ready gate.

### 6.10 Joint classical--PFHE interpretation

Classical metrics and PFHE answer different questions about the same outputs. The classical view measures effect magnitude in the metric defined by each task. PFHE measures the composition of that effect across paired evaluation units. For binary correctness, the two are linked exactly by `DeltaAcc=(A-H)/N`, but PFHE additionally distinguishes recovery, regression risk, and the mass of outcomes that remain unchanged. For partial-credit tasks, the sign and magnitude of native-score changes remain separate from strict correctness transitions; neither view can be inferred from the other.

Tables 13--17 instantiate this logic as exploratory PFHE-format record-level analyses under their stated historical scoring protocols. Their endpoints are paired score transitions and native magnitudes; operational endpoints require prospective evaluation.

The classical results therefore remain the primary empirical account: Qwen3.5 has positive exploratory tool-route selector effects; Qwen3 and Qwen2.5 retain source-specific oracle headroom without a positive train-selected fixed tool route; and Qwen3.5 NS versus non-NS behavior varies by task. PFHE does not replace those comparisons or create a second ranking of models. It asks whether a classical effect is supported by more fixes than harms, whether regression risk is acceptable, and how much of the paired universe is preserved.

This joint reading prevents two symmetric overclaims. A positive native-score delta does not by itself show that strict harms are controlled. Conversely, a positive fix-minus-harm count does not imply that mean partial-credit magnitude is positive. Evidence strata remain separate: a combination-oracle PFHE-format row describes candidate-space headroom, a fixed-route row describes one frozen intervention, and a learned-selector row describes only the selected outputs admitted by that selector's feature and split contract.

The RF comparison illustrates the distinction between endpoints and scope. Its native effect, strict four-cell balance, and paired RF--linear contrast are favorable within the declared six-task partition.

Their agreement does not validate feature availability at decision time, an acceptance threshold, baseline fallback, or transfer to new partitions and checkpoints.

## 7. Prospective Deployment Requirements

The deployment implication is intentionally narrow and prospective. The current experiments do not validate deployment because the learned and RF selectors force an AG choice and do not execute acceptance or baseline fallback. Qwen3.5 function-call/tool routes are candidates for prospective validation; the source-separated Qwen3 and Qwen2.5 results do not establish transferable routing.

We define three prospective validation profiles rather than validated deployment recipes.

**Table 26. Prospective validation profiles for future AG acceptance-and-fallback studies.**

| Profile | Purpose | Gate style | Use when |
|---|---|---|---|
| Safety-first | Minimize harm | Conservative low-coverage gate and baseline fallback | High-risk domains |
| Balanced | Trade off effect and regression risk | Validated category/model-specific selector | General tool calling |
| Performance-first | Capture more candidate space | Higher-coverage accepted route with monitoring | Internal or batch workflows |

We do not recommend multi-flow multi-candidate voting as a first production architecture. It is useful offline, but online cost scales with the number of candidates. A more realistic path is a single-flow or lightly augmented baseline-plus-candidate design with online feature logging and fallback.

The current engineering evidence is neither a deployment estimate nor a serving benchmark. Hidden-dimensional points such as P1s and P3 require less activation storage than expansion-dimensional P2/P2a/P2b points, while offline multi-candidate mining is far more expensive than the intended online route. In the Qwen3.5-9B ToolBench audit, gate features such as early trajectory and cross-PPL materially reduce harm exposure, but post-fork cross-PPL is diagnostic-only in the present implementation because it depends on generated candidate tokens. A production version should therefore either use online approximations of those signals or measure the cost of a lightly augmented candidate path directly. We do not claim that AG is cost-free; the claim is narrower, namely that single-candidate hidden-dimensional routes are the plausible first deployment target, while broad multi-candidate voting belongs in offline mining.

The first prospective serving tests should compare hidden-dimensional points such as P1s or P3 against intermediate-dimensional points before any preference is adopted. P2/P2a/P2b remain scientifically important, but their deployment cost and noise profile require direct measurement.

## 8. Limitations

This paper makes a deliberately bounded claim. The model evidence is limited to Qwen-family checkpoints. Qwen3.5-9B, Qwen3-8B, and Qwen2.5-7B cover useful attention and FFN-shape variation, while Qwen3.5 supplies the aligned 24-candidate historical union. Qwen3 old-P/new-P comparisons use distinct sample domains and baseline outcomes; Qwen2.5 sources also differ in baseline outcomes. The interpretation is one positive within-checkpoint exploratory selector result plus source-separated cross-checkpoint boundary evidence, rather than universal model transfer.

The task evidence is concentrated. The strongest exploratory selector gains appear on Qwen3.5 tool/function-call routes. Financial QA is represented by one endpoint, and multiple-choice diagnostics are weak or negative. Category micro-averages combine heterogeneous native scorers and remain secondary to dataset-level endpoints.

The selector evidence concerns forced-choice offline ranking. The learned and RF selectors use baseline as an evaluation reference. Trajectory-derived features are analyzed offline; prospective evaluation must establish feature timing and serving cost and test acceptance and fallback. RF results use one grouped five-fold partition on one model. Positive paired intervals condition on its fitted policies and do not establish stability across refits, independent partitions, or checkpoints.

PFHE is interpreted within its pairing and scorer contracts. Strict full-credit transitions are uninformative on some partial-credit datasets, where native magnitude remains primary. The reported PFHE-format analyses characterize paired saved-score records; prospective operational evaluation should measure timeout and error status separately. Historical raw 0.1-change and decile-crossing `partial10` results retain their separate estimands and are excluded from confirmatory PFHE comparisons.

Mechanistically, the P1s/P2a/P2b refinement motivates controlled tests of gate-branch correction and branch-specific effects. Current reports show that these SwiGLU points can help, while first-divergence and autoregressive- propagation diagnostics characterize how a local intervention can affect later tokens. Layer-trace outcome aggregates are descriptive mechanism probes, rather than evidence for a standalone layer gate or a causal low/mid/high-layer optimum. Layer-band ablations and direct tests of late-layer amplification remain future work.

Finally, all selection evidence remains offline. Repeated grouped splits, time-split and prospective validation, calibration of acceptance thresholds, and serving measurements of latency, throughput, and memory are still required.

## 9. Future Work: Validation, Statistics, and Mechanistic Tests

Several validation steps remain before stronger deployment claims are justified.

Prospective evaluation should extend these protocol-specific analyses to independent checkpoints and datasets, preserving fixed sample universes, scorer versions, and request-status provenance. Such studies would assess transfer and stability under operational conditions.

Evaluation of RF beyond the current experimental conditions should use grouped outer test splits and grouped inner model/feature selection, with all preprocessing confined to training folds. Neutral candidates and the baseline must be explicit actions so that the model learns acceptance or abstention rather than forced ranking. Repeated grouped splits, paired RF--linear intervals, independent datasets, and cross-model tests are required. Stacking should be revisited only with genuine out-of-fold meta-features and a separately held-out final test set.

PFHE also requires broader validation. Future work should use dataset-aware paired resampling, report requested/successful/paired/error counts, test scorer and record-status sensitivity, and measure stability across decoding seeds. For partial-credit tasks, a minimally important difference should be declared before converting score magnitudes into transition-style sensitivity analyses. Prospective studies should pre-register acceptable regression-risk and coverage regions and report classical and PFHE endpoints on the same frozen universe.

Mechanistic validation should test the selection-signal hypothesis in Section 6.1 by transferring identical masks between expanded-channel sites, then comparing adaptive masks at matched retained-coordinate budgets. Window size and NS/non-NS choice also require one aligned protocol.

**Joint P2a+P2b intervention.** We will test complementary gate- and value-derived masks within the same SwiGLU layer. With binary masks on both branches, a fused channel survives only if both masks retain it; the effective mask is their intersection. Joint masking may suppress more channels without improving selection, so it must be compared with each single-site variant at matched effective sparsity and compute. Independently computed branch masks will be compared with a coupled rule that conditions value selection on the gate mask and fused signal. Held-out task scores, paired fixes/harms, and serving cost will determine whether combining the signals improves on P2b alone.

**Combining multiplicative and additive interventions.** Future work will explore coordinate-gated scaling together with additive steering, for example `x'=D(x,history)x+βv`. Adding the offset before gating may change both the adaptive mask and which parts of the steering signal survive, so intervention order requires evaluation. Equal-budget tests should compare additive-only, multiplicative-only, and combined variants under matched sites and decoding protocols, reporting task scores, paired fixes and harms, latency, throughput, and memory. The combined operation is a future research direction.

**Engineering AG for efficient serving.** Production use must address the latency, throughput, and memory costs of AG's additional activation statistics, mask construction, and execution. The engineering case discussed in the referenced article [24] illustrates the potential of specialized inference: Baseten reports that VibeQwen, running Qwen-3.6-35B-A3B in NVFP4 on one B200, achieved 1.90 times the single-stream decoding throughput of tuned vLLM 0.25.1 on repetitive, structured text, and 1.71 times its output throughput at concurrency 32 [23]. The engines were still experimental rather than serving production traffic. This is evidence of optimization headroom in a different system, not a measured AG speedup. A comparable profile--optimize--verify process could substantially reduce, and potentially offset, AG's added cost. For AG, we propose model-, hardware-, and workload-specific kernels, fusion of statistics reductions and amplitude masking, fewer kernel launches and intermediate-memory transfers, and graph capture or batching where compatible with the activation-history contract. These are future engineering proposals; the intervention site, mask definition, retained-mass rule, and decode phase must remain intact. Evaluation must optimize both baseline and AG paths under the same hardware, precision, batch sizes, sequence lengths, and traffic, then report incremental TTFT, time per output token, throughput, peak memory, and tail latency alongside output-quality and numerical checks. Only that comparison can establish whether engineering has compensated AG's performance degradation while retaining its task benefits.

Training-time use of FFN signals is another open direction. This paper studies the inference-time form of FFN intervention: AG creates candidate forward paths, and the proposed controller would accept one only when benefit is predicted. The same FFN-level signals may also be useful before deployment, for example to identify noisy tool-supervision examples, detect unstable schema or argument annotations, construct activation-aware curricula, or train gates with cleaner positive and negative examples. If FFN trajectories reveal where a model's structured decision becomes unstable, they may help improve both inference-time routing and the quality of the data used to train tool-structured behavior. We leave this training-side extension to future work.

## 10. Conclusion

Amplitude Gating tests coordinatewise FFN masking while keeping pretrained weights fixed. The six SwiGLU sites separate input coverage, expanded-channel selection signals, and the output update; the non-NS/NS comparison changes coordinate ranking under a common entropy-mass rule. The ORP failures motivate this alternative operation without proving its safety or causal mechanism.

The corrected September Qwen3.5 fixed routes show task-dependent benefits. The two function-call and two ToolBench non-NS effects remain positive when the weaker observed AG result is compared with the stronger observed baseline. JSON gains are small. These are local lexical or answer/action-name scores, with robustness established only against the two observed executions. Population significance, future-run stability, executed-tool success, and equal-compute advantages require further evaluation.

The historical aligned Qwen3.5 offline selector improves the descriptive tool-route micro-average by +4.27 pp and the two function-call endpoints by about +7.6 pp. Its same-output PFHE-format view records 48 fixes and 26 harms over 2,556 units. These estimates characterize the source-aligned historical protocol and are reported separately from the September fixed-route results. Qwen3 and Qwen2.5 retain source-specific oracle headroom without a positive train-selected fixed tool route. Candidate availability therefore cannot substitute for validated selection.

PFHE makes paired benefit and risk visible alongside task scores. Under the current experimental conditions, the six-task RF refit has a +0.46 pp native effect and +0.57 pp paired advantage over the linear ranker, with positive conditional intervals in one grouped partition. ConvFinQA-only mixed and ridge refits give small absolute gains on 533 held-out units. These findings remain offline forced-selection evidence. Stable transfer, acceptance, baseline fallback, feature timing, and serving cost require prospective evaluation.

## Appendix A. Supplementary scores and data provenance

The hand-eight-shot GSM8K rows below share one prompt protocol and their own baseline. These are fixed raw measurements, not held-out selector estimates. No winner is selected from them to replace the separately reported historical GSM8K Table 11 summary. The SVD-ORP shared-P1 row is the corrected full-domain diagnostic.

**Table A1. September supplementary native scores under the hand-eight-shot GSM8K protocol.**

| Protocol | Configuration | N | Native score | Delta vs own baseline |
|---|---|---|---|---|
| GSM8K hand-8-shot | baseline | 1319 | 87.9454% | +0.0000 pp |
| GSM8K hand-8-shot | ns_entropy_ag P1 e0.93 w4 | 1319 | 87.7180% | -0.2274 pp |
| GSM8K hand-8-shot | ns_entropy_ag P1 e0.95 w4 | 1319 | 87.4905% | -0.4549 pp |
| GSM8K hand-8-shot | ns_entropy_ag P1s e0.93 w4 | 1319 | 88.9310% | +0.9856 pp |
| GSM8K hand-8-shot | ns_entropy_ag P1s e0.95 w4 | 1319 | 87.9454% | +0.0000 pp |
| GSM8K hand-8-shot | ns_entropy_ag P1v e0.93 w4 | 1319 | 86.9598% | -0.9856 pp |
| GSM8K hand-8-shot | ns_entropy_ag P1v e0.95 w4 | 1319 | 87.3389% | -0.6065 pp |
| GSM8K hand-8-shot | ns_entropy_ag P2 e0.93 w4 | 1319 | 87.4147% | -0.5307 pp |
| GSM8K hand-8-shot | ns_entropy_ag P2 e0.95 w4 | 1319 | 87.4147% | -0.5307 pp |
| GSM8K hand-8-shot | pure_entropy_ag P1 e0.93 | 1319 | 87.4905% | -0.4549 pp |
| GSM8K hand-8-shot | pure_entropy_ag P1 e0.95 | 1319 | 88.5519% | +0.6065 pp |
| GSM8K hand-8-shot | pure_entropy_ag P1s e0.93 | 1319 | 88.0970% | +0.1516 pp |
| GSM8K hand-8-shot | pure_entropy_ag P1s e0.95 | 1319 | 88.0970% | +0.1516 pp |
| GSM8K hand-8-shot | pure_entropy_ag P1v e0.93 | 1319 | 88.0970% | +0.1516 pp |
| GSM8K hand-8-shot | pure_entropy_ag P1v e0.95 | 1319 | 88.1729% | +0.2274 pp |
| GSM8K hand-8-shot | pure_entropy_ag P2 e0.93 | 1319 | 88.4761% | +0.5307 pp |
| GSM8K hand-8-shot | pure_entropy_ag P2 e0.95 | 1319 | 87.9454% | +0.0000 pp |
| GSM8K hand-8-shot | svd_orp P1 e0.95 w3 | 1319 | 8.4155% | -79.5299 pp |

The released configuration index binds retained measurements to their original prompts, model weights, scorers, and sample universes. Current file checksums and repository-relative source paths are recorded in the release manifest.

Tables 9--10, 13--14, 16--18, 22--25 and Figures 7, 10 and 11 report the archived source-aligned or source-separated analyses. Tables 10a and 19 report the single-task financial-QA and six-task RF refits, respectively. Tables 11--12, 15, 20--21, 22a and A1, and Figures 1 and 9 report September measurements or explicitly marked identity controls. Figures 3 and 4 show execution timing and physical sites; Figure 8 visualizes Tables 11 and 11a. Each result is interpreted within its stated protocol.

**Tool use.** OpenAI Codex was used to assist with language editing and figure preparation. The authors are responsible for the scientific claims, data, and citations.